\documentclass[lettersize,journal]{IEEEtran}
\usepackage{amsmath,amsfonts}
\usepackage[ruled,linesnumbered]{algorithm2e} 
\usepackage{array}
\usepackage[caption=false,font=normalsize,labelfont=sf,textfont=sf]{subfig}
\usepackage{textcomp}
\usepackage{stfloats}
\usepackage{url}
\usepackage{foreign}
\usepackage{verbatim}
\usepackage{booktabs}
\usepackage{multirow}
\usepackage{graphicx}
\usepackage{graphicx}
\usepackage{xcolor}

\usepackage{cite}
\begin{document}

\title{Towards Transferable Defense Against Malicious Image Edits}

\author{Jie Zhang,~\IEEEmembership{Member,~IEEE,}
\and
Shuai Dong,
\and
Shiguang Shan~\IEEEmembership{Fellow,~IEEE,}
\and
Xilin Chen~\IEEEmembership{Fellow,~IEEE,}
\thanks{This work is partially supported by Strategic Priority Research Program of the Chinese Academy of Sciences (No. XDB0680202),
Beijing Nova Program (20230484368), and Youth Innovation Promotion Association CAS. Corresponding author: Shiguang Shan.}
\thanks{Jie Zhang, Shiguang Shan and Xilin Chen are with the State Key Laboratory of AI Safety, Institute of Computing Technology, Chinese Academy of Sciences (CAS), Beijing 100190, China, and also with the University of China Academy of Sciences, Beijing 100049, China (e-mail: zhangjie@ict.ac.cn; sgshan@ict.ac.cn; xlchen@ict.ac.cn). }
\thanks{Shuai Dong is with the School of Computer Science, China University of Geosciences, Wuhan 430074, China (e-mail: dongshuai\_iu@cug.edu.cn).}
}
\markboth{Journal of \LaTeX\ Class Files,~Vol.~14, No.~8, August~2021}%
{Shell \MakeLowercase{\textit{et al.}}: A Sample Article Using IEEEtran.cls for IEEE Journals}

\IEEEpubid{0000--0000/00\$00.00~\copyright~2021 IEEE}

\maketitle

\begin{abstract}
Recent approaches employing imperceptible perturbations in input images have demonstrated promising potential to counter malicious manipulations in diffusion-based image editing systems. However, existing methods suffer from limited transferability in cross-model evaluations. To address this, we propose  \textbf{T}ransferable \textbf{D}efense \textbf{A}gainst Malicious Image \textbf{E}dits (TDAE), a novel bimodal framework that enhances image immunity against malicious edits through coordinated image-text optimization. Specifically, at the visual defense level, we introduce FlatGrad Defense Mechanism (FDM), which incorporates gradient regularization into the adversarial objective. By explicitly steering the perturbations toward flat minima, FDM amplifies immune robustness against unseen editing models. For textual enhancement protection, we propose an adversarial optimization paradigm named Dynamic Prompt Defense (DPD), which periodically refines text embeddings to align the editing outcomes of immunized images with those of the original images, then updates the images under optimized embeddings. Through iterative adversarial updates to diverse embeddings, DPD enforces the generation of immunized images that seek a broader set of immunity-enhancing features, thereby achieving cross-model transferability. Extensive experimental results demonstrate that our TDAE achieves state-of-the-art performance in mitigating malicious edits under both intra- and cross-model evaluations.
\end{abstract}


\section{Introduction}
\label{sec:intro}
\IEEEPARstart{D}{iffusion}-based generative models, which have their theoretical underpinnings in foundational works such as \cite{Ho2020DenoisingDP,SohlDickstein2015DeepUL,Song2020ScoreBasedGM}, have precipitated significant advancements in the domain of digital image synthesis and manipulation. Subsequent research endeavors have substantially augmented these capabilities, most notably through the sophisticated integration of natural language guidance, as evidenced by a multitude of influential studies \cite{Avrahami2021BlendedDF,Couairon2022DiffEditDS,Hertz2022PrompttoPromptIE,Meng2021SDEditGI,Mokady2022NulltextIF,Rombach2021HighResolutionIS,Saharia2022PhotorealisticTD,Wallace2022EDICTED,Xie2022SmartBrushTA,Xu2023OpenVocabularyPS,Zhu2023MovieFactoryAM}. Moreover, this powerful generative paradigm is no longer confined to static images, with recent breakthroughs extending these fine-grained editing capabilities to the dynamic domain of video, enabling sophisticated motion transfer and high-fidelity animation \cite{Tu2023MotionEditorEV, Tu2024StableAnimatorHI}. This synergistic interplay between visual generation processes and nuanced textual control empowers users to create and edit media with unprecedented levels of fidelity and semantic coherence. While this technological progress enables diverse applications, \textit{e.g.}, from innovative creative expression and scientific visualization to personalized content, it also introduces complex ethical dilemmas and significant societal risks. Malicious actors can readily exploit the ability to create highly convincing fake photorealistic content for harmful purposes \cite{Maras2018DeterminingAO,Pei2024DeepfakeGA}. Synthetic media can be weaponized to spread misinformation, violate privacy through non-consensual imagery, or manipulate public discourse. The unchecked spread of easily created, persuasive fake images threatens to erode trust in digital media and destabilize socio-political systems. Therefore, exploring security measures for image editing technology is essential to mitigate these risks.
\IEEEpubidadjcol
In response to these escalating risks, the research community has principally pursued two distinct mitigation paradigms: the reactive detection of manipulated content \cite{Passos2022ARO,Naitali2023DeepfakeAG,Pei2024DeepfakeGA,wang2022lisiam,han2023fcd,yin2023dynamic,10061274,10411047,9468380,10209264} and the proactive immunization of pristine images against unauthorized modifications \cite{Aneja2021TAFIMTA,Ruiz2020DisruptingDA,Ruiz2023PracticalDO,Yeh2021AttackAT,Salman2023RaisingTC,Xue2023TowardEP}. Reactive detection methods operate after the fact, using classifiers to identify digital artifacts or semantic inconsistencies that signal manipulation. While valuable for digital forensics, this approach inherently cedes the initiative: malicious content can spread widely and cause harm before it's detected. Conversely, proactive immunization establishes pre-emptive defenses by embedding engineered, often imperceptible adversarial perturbations into source images before distribution. This approach’s key advantage is disrupting malicious editing at inception rather than detecting it post-creation. By shifting security efforts from downstream analysis to the content origin point, it empowers creators with direct control over digital assets.

\begin{figure*}[t]
    \centering
    \includegraphics[width=1\linewidth]{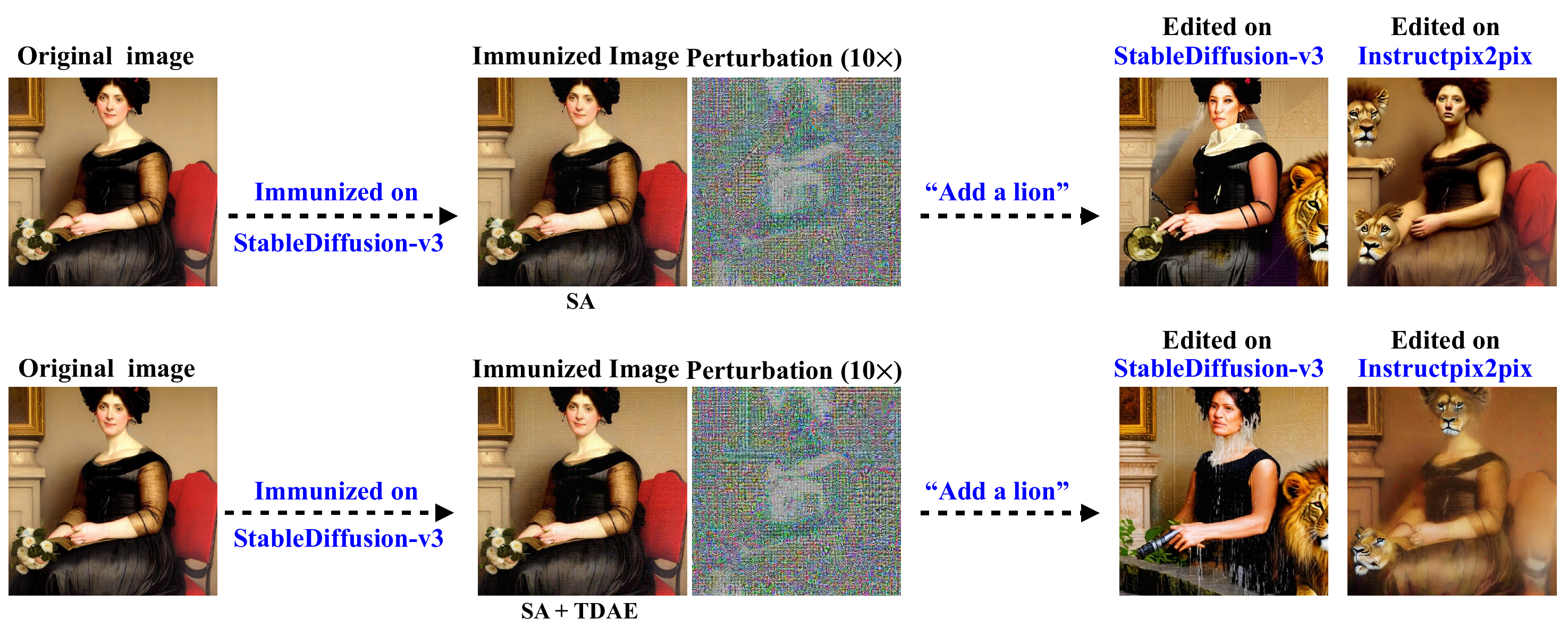}
    \caption{Qualitative comparison demonstrating TDAE's enhanced cross-model transferability. Both original images are immunized on StableDiffusion v3 to resist the edit prompt ``Add a lion''. The top row utilizes the SA~\cite{Lo_2024_CVPR} method, while the bottom row employs SA+TDAE. Although the SA method shows limited effectiveness when transferred to the unseen InstructPix2Pix model, SA+TDAE successfully disrupts the malicious edit on both the source or surrogate model, StableDiffusion v3, and the unseen model, InstructPix2Pix, showcasing superior generalization.}
    \label{fig:fig1}
\end{figure*}

Early immunization techniques, primarily developed for GAN-based editing models \cite{Aneja2021TAFIMTA,Ruiz2020DisruptingDA}, have exhibited initial efficacy. However, the intrinsic robustness and advanced denoising capabilities inherent to diffusion models introduce novel challenges, thereby compelling the development of more specialized defensive strategies. Subsequent research has consequently concentrated on creating tailored defenses specifically for diffusion models, investigating a diverse array of attack vectors. Initial methodologies for these models involve adversarial attacks targeting either the Variational Autoencoder (VAE) component or the entirety of the image generation pipeline, exemplified by approaches such as PGE and PGD \cite{Salman2023RaisingTC}. This foundational work has since evolved into more sophisticated strategies, including disrupting semantic comprehension by perturbing cross-attention mechanisms \cite{Lo_2024_CVPR}. Another avenue involves preventing unauthorized style replication through the integration of textural loss functions, a technique employed by Mist \cite{Liang2023MistTI}. Moreover, considerable effort has been dedicated to augmenting the efficiency and effectiveness of these adversarial attacks. One such method is score distillation sampling \cite{Xue2023TowardEP}, a plug-and-play strategy that improves efficiency by approximating the gradient calculation. Another approach is the targeted attack \cite{Zheng2023TargetedAI}, which works by minimizing the distance between the editing model's output and a pre-defined target. Despite significant advances and diverse methodologies, a core limitation persists: adversarial perturbations are typically generated through attacks on, and are optimized for, specific surrogate models. Consequently, while immunized images demonstrate strong defense against the models used to generate their perturbations, their protection degrades substantially against unseen editing models, as shown in Fig.~\ref{fig:fig1}.

In this paper, we propose \textbf{T}ransferable \textbf{D}efense \textbf{A}gainst Malicious Image \textbf{E}dits (TDAE), a novel bimodal framework that enhances adversarial robustness by synergistically optimizing
both image and text modalities. Unlike traditional methods, TDAE specifically enhances the transferability of immunized images, which consists of two key parts: \textbf{F}latGrad \textbf{D}efense \textbf{M}echanism (FDM) and \textbf{D}ynamic \textbf{P}rompt \textbf{D}efense (DPD). Inspired by~\cite{Fan2023TransferabilityBT}, we propose FDM by minimizing the expectation of gradient norms across perturbed images at the image level for enhancing immunized image transferability. FDM replaces the conventional expectation minimization with direct optimization of the maximum gradient norm under local perturbations, eliminating sampling requirements while maintaining robust defense effectiveness against unseen models. This reformulation achieves substantial efficiency gains without sacrificing effectiveness. Complementing FDM, we introduce DPD at the text level, leveraging insights from multi-prompt optimization~\cite{Luo2024AnII}. DPD employs an iterative process to refine text embeddings, ensuring that edits on immunized images align with those of the original images. Concurrently, it updates immunized images using these optimized text embeddings, creating a dual-loop optimization that generates immunized images by seeking a broader set of immunity-enhancing features. Broader features bring potential transferability to unseen models, leading to better cross-model immunity effectiveness. It should be noted  that our TDAE is a plug-and-play module which can be integrated into any image immunization method for improving cross-model transferability. In summary, we make the following contributions:

\begin{itemize}
    \item We propose \textbf{T}ransferable \textbf{D}efense \textbf{A}gainst Malicious Image \textbf{E}dits (TDAE), a novel framework designed to enhance the cross-model transferability in image immunization. This approach enables immunized images to maintain robust protection even when transferred across unseen editing models.
    \item The framework introduces two synergistic mechanisms: an image-level FlatGrad Defense Mechanism and a text-level Dynamic Prompt Defense. Through a coordinated bi-level optimization strategy, TDAE generates immunized images with flat minima (FDM) while incorporating more immunity-enhancing features (DPD). This synergistic interaction substantially improves sample robustness against unseen malicious edits.
    \item Extensive experiments show that TDAE achieves the superior immunity effectiveness in both intra-model and cross-model scenarios. 
\end{itemize}

\section{Related Work}
\label{sec:related_work}
This section reviews literature related to our proposed TDAE, covering three primary areas. These include advancements in diffusion-model-based image editing, image immunization techniques, and methodologies for enhancing the transferability of adversarial attacks.

\subsection{Image Editing}
Image editing aims to semantically modify a source image according to textual prompts, ensuring the output aligns with instructions while preserving visual coherence. The ascendancy of diffusion models has catalyzed remarkable progress in this area, offering unprecedented control and realism. Early techniques like SDEdit \cite{meng2021sdedit} employ stochastic differential equations for iterative, text-guided image refinement through a noise-and-denoise process. Blended Diffusion \cite{avrahami2022blended} advances this with localized, text-guided edits that preserve unedited regions. Subsequent methods leverage vision-language models such as CLIP \cite{Radford2021LearningTV}. DiffusionCLIP \cite{kim2021diffusionclip}, for instance, uses CLIP embeddings for text-guided manipulation, enabling zero-shot editing without paired training data. For finer-grained control, Prompt-to-Prompt \cite{hertz2022prompt} enables sophisticated semantic alterations by manipulating textual prompts to steer cross-attention maps, which link text to visual regions. ControlNet \cite{zhang2023controlnet} achieves explicit spatial and structural control by conditioning generation on auxiliary inputs like edge maps or segmentation masks, ensuring adherence to precise spatial constraints. These advancements represent the forefront of text-guided image editing, offering potent tools whose potential for misuse warrants careful consideration. Besides, this powerful generative paradigm is no longer confined to static images, with recent breakthroughs extending these fine-grained editing capabilities to the dynamic domain of video, enabling sophisticated motion transfer and high-fidelity animation \cite{Tu2023MotionEditorEV, Tu2024StableAnimatorHI}. Specifically, MotionEditor \cite{Tu2023MotionEditorEV} incorporates a novel content-aware motion adapter to manipulate motion information while rigorously preserving the original protagonist’s appearance and background. Concurrently, StableAnimator \cite{Tu2024StableAnimatorHI} achieves state-of-the-art identity-preserving human animation by employing a distribution-aware ID adapter and specific optimization strategies to synthesize high-quality videos without reliance on post-processing. These cutting-edge advancements in generative editing provide powerful tools that, while innovative, pose significant risks through their potential to create convincing fake media.

\subsection{Image Immunization}
The rapid evolution of diffusion-based generative models has significantly lowered the barrier to malicious image manipulation, thereby accentuating the urgent need for robust countermeasures. Existing image immunization techniques address this challenge by strategically embedding targeted, often imperceptible, adversarial perturbations into original images \cite{Shan2020FawkesPP,Ruiz2020DisruptingDA,Aneja2021TAFIMTA,Dong2017DiscoveringAE,MoosaviDezfooli2015DeepFoolAS,Xie2017AdversarialEF} to disrupt unauthorized modifications. For diffusion-based image editing models, Salman \textit{et al.} \cite{Salman2023RaisingTC} introduce two pioneering approaches named PGE and PGD. The PGE \cite{Salman2023RaisingTC} method perturbs the VAE's latent distribution, offering simplicity but limited protection, while PGD \cite{Salman2023RaisingTC} disrupts the entire forward denoising process for stronger defense at a high computational cost. Building on these initial efforts, subsequent research explores more nuanced strategies. To mitigate high computational overheads, Lo \textit{et al.} \cite{Lo_2024_CVPR} propose selectively targeting text-salient regions via cross-attention layers. Concurrently, Mist \textit{et al.} \cite{Liang2023MistTI} incorporate textural losses to prevent unauthorized style mimicry by minimizing the distance in the latent space between the protected image and a pre-defined target image. Further efforts aim at enhancing attack effectiveness and efficiency. For example, Xue \textit{et al.} \cite{Xue2023TowardEP} leverage score distillation sampling, a plug-and-play strategy that improves efficiency by approximating the gradient calculation. Zheng \textit{et al.} \cite{Zheng2023TargetedAI} investigate targeted attack, which works by minimizing the distance between the editing model's output and a pre-defined target. Despite significant progress, a key challenge persists: images immunized against specific models often lose effectiveness against unseen or architecturally different models. This underscores the critical need for more generalized, model-agnostic immunization strategies.

\subsection{Transferability of Adversarial Attacks}
\label{trans} 
In practice, enhancing the transferability of adversarial images from surrogate to unseen black-box models is crucial, as many models are deployed without direct white-box access. Initial strategies include data augmentation techniques~\cite{Wang2021AdmixET,Xie2018ImprovingTO} like scaling and input translation~\cite{Lin2019NesterovAG}, and optimization-based methods~\cite{Dong2017BoostingAA,Lin2019NesterovAG} using advanced optimizers with momentum. More recently, approaches leveraging the geometry of the loss landscape have shown significant promise for improving transferability by seeking flat minima. Notable examples include RAP by Qin \textit{et al.}~\cite{NEURIPS2022_c0f9419c}, which introduces a \textit{reverse} adversarial perturbation during the attack process to push the solution away from sharp minima and toward flatter regions of the loss landscape. Fan \textit{et al.}~\cite{Fan2023TransferabilityBT} propose TPA, which achieves state-of-the-art adversarial transferability enhancement results by minimizing the expected gradient norm within a neighborhood of the perturbed image. While these methods have advanced adversarial transferability, their primary application has been towards \textit{discriminative models} like classifiers. The specific challenge of enhancing such transferability for adversarial immunizations against \textit{generative models}, particularly diffusion-based image editing systems, remains less explored. Drawing inspiration from TPA, our FlatGrad Defense Mechanism (FDM), a core component of TDAE, is designed to improve immunization transferability against these generative systems. Although TPA is effective, it incurs significant computational overhead from multi-sample averaging. In contrast, FDM refines this by substituting expectation minimization with the direct optimization of the \textit{maximum gradient norm} across local perturbations, which is further approximated for efficiency as detailed in Section \ref{fdm}. This reformulation achieves substantial efficiency gains while maintaining robust immunization effectiveness and transferability, offering a more practical solution for immunizing against generative editing models.

\section{Method}
\subsection{Preliminaries}
\label{sec:preliminaries} 
Let a source image be denoted by \( x_0 \in \mathcal{X} \) and a textual editing embedding by \( c \in \mathcal{C} \). We denote the desired edited output as \( y_0 \in \mathcal{Y} \). The image editing system, \( f_\theta: \mathcal{X} \times \mathcal{C} \rightarrow \mathcal{Y} \), parameterized by \( \theta \), is tasked with generating outputs such that \( f_\theta(x_0, c) \approx y_0 \) with high fidelity. We define an \( \epsilon \)-neighborhood around \( x_0 \) as \( \mathcal{B}_\epsilon(x_0) = \{ x' \in \mathcal{X} \mid \|x' - x_0\|_p \leq \epsilon \} \). Here, \( \epsilon > 0 \) specifies the permissible perturbation magnitude, and \( \| \cdot \|_p \) denotes the \( L_p \)-norm, with \( L_\infty \) being commonly used for imperceptibility.

The objective of image immunization is to compute an optimal perturbation \( \delta_v^* \) from a set of all possible perturbations \( \delta_v \). The immunized example \( x_{\text{adv}} = x_0 + \delta_v \) must satisfy two key conditions. First, the perturbation must be constrained such that \( \|\delta_v\|_p \leq \epsilon \), implying \( x_{\text{adv}} \in \mathcal{B}_\epsilon(x_0) \). Second, the output of the image editing model, \( f_\theta(x_{\text{adv}}, c) \), should exhibit maximal deviation from the desired edited output \( y_0 \).

Formally, let \( \mathcal{L}: \mathcal{Y} \times \mathcal{Y} \rightarrow \mathbb{R}_{\ge 0} \) denote a differentiable loss metric (e.g., \( \ell_2 \)-distance) quantifying the discrepancy between model outputs and target results. As established in adversarial attack literature \cite{Goodfellow2014ExplainingAH,Kurakin2016AdversarialEI}, this objective can be formulated as a constrained maximization problem:
\begin{equation}
\delta_v^* = \underset{\|\delta_v\|_p \leq \epsilon}{\arg\max}\, \mathcal{L}\left( f_\theta(x_0 + \delta_v, c), y_0 \right).
\label{eq:adv_objective}
\end{equation}

These immunization methods primarily generate immunized examples through gradient-based optimization techniques. A prominent example is the Projected Gradient Descent algorithm \cite{Madry2017TowardsDL}. Projected Gradient Descent iteratively updates the adversarial perturbation \( \delta_v \) to maximize the loss function while ensuring the perturbation remains within the \( \epsilon \)-neighborhood constraint. Specifically, starting from an initial perturbation (e.g., \( \delta_{v0} = 0 \)), at each iteration $t$, the perturbation \( \delta_{vt} \) is updated as follows:
\begin{equation}
    \delta_{v(t+1)} = \Pi_{\|\cdot\|_p \leq \epsilon} \left( \delta_{vt} + \alpha \, \text{sign}\left(\nabla_{\delta_{vt}} \mathcal{L}(f_\theta(x_0 + \delta_{vt},c), y_0)\right) \right),
\label{eq:pgd_update} 
\end{equation}
where $\alpha > 0$ is the step size, $\nabla_{\delta_{vt}} \mathcal{L}(f_\theta(x_0 + \delta_{vt},c), y_0)$ is the gradient of the loss with respect to the perturbation $\delta_{vt}$ at iteration $t$, and $\Pi_{\|\cdot\|_p \leq \epsilon}$ denotes the projection operator that ensures the updated perturbation satisfies \( \|\delta_{v(t+1)}\|_p \leq \epsilon \). The \(\text{sign}(\cdot)\) function ensures updates follow the direction of steepest ascent. Projected Gradient Descent thus iteratively maximizes the loss while constraining perturbations.

\begin{figure*}[t]
    \centering
    \includegraphics[width=0.9\linewidth]{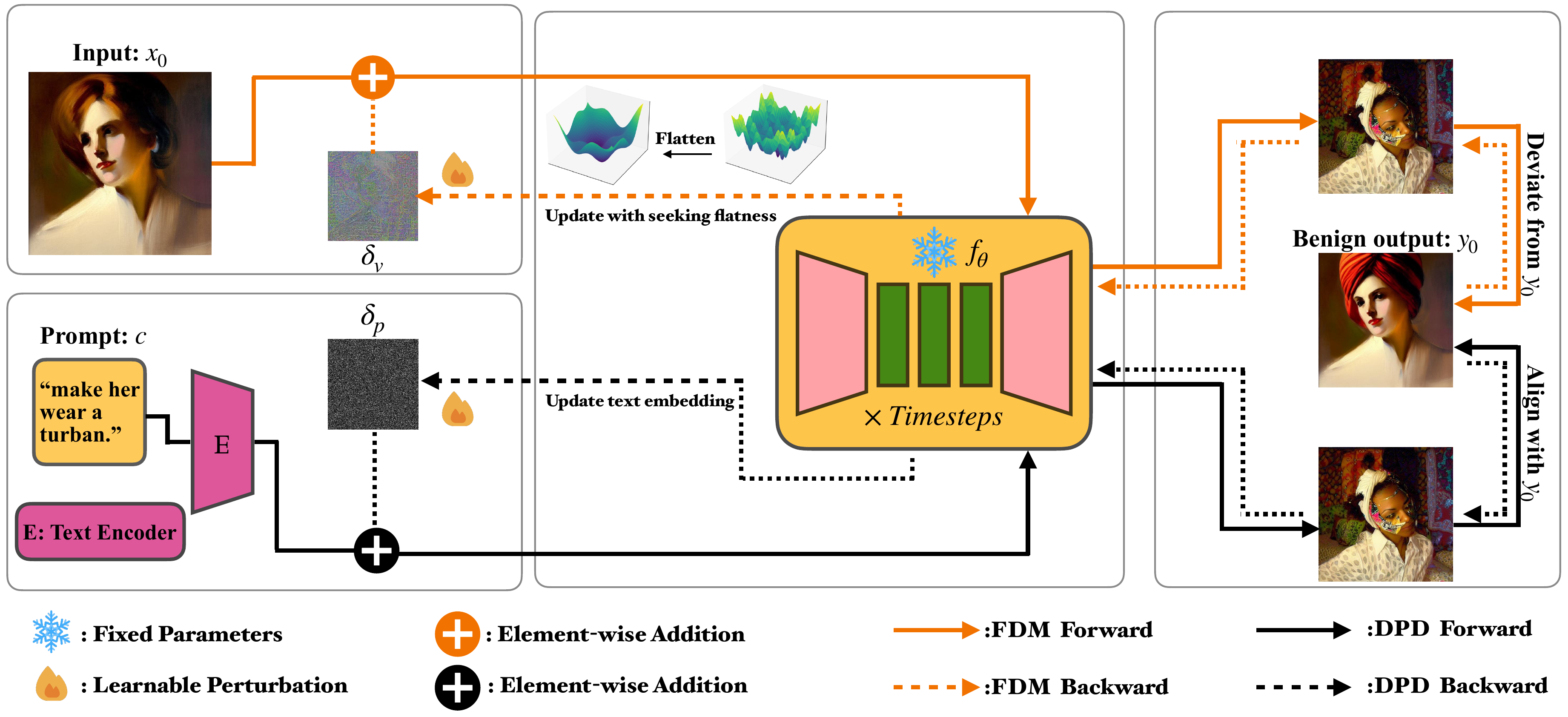}
    \caption{The Overview of Our TDAE. TDAE consists of two key parts, \textit{i.e.}, FlatGrad Defense Mechanism (FDM) and Dynamic Prompt Defense (DPD), which operates through coordinated optimization of two trainable perturbations: image perturbation \(\delta_v\) and text-embedding perturbation \(\delta_p\). The visual defense component \(\delta_v\) is adversarially optimized to maximize the discrepancy between editing results of the immunized image and the original (benign) image outputs, while adhering to gradient norm constraints to ensure convergence to flat minima. Complementarily, the textual defense component \(\delta_p\) is iteratively refined to minimize this discrepancy, aligning editing outcomes of the immunized image with benign results. To synchronize these dual defenses, every S optimization steps, \(\delta_v\) undergoes re-optimization using the updated text embedding (\(c + \delta_p\)), where \(c\) denotes the original text embedding. This cyclical adversarial process reinforces cross-modal alignment, generating the immunized image by seeking robust, transferable features against diverse editing models.}
    \label{fig:framework}
\end{figure*}
\subsection{Overall Framework}
Our proposed Transferable Defense Against Malicious Image Edits (TDAE) framework is designed to overcome the critical limitations of existing image immunization methods, particularly their lack of cross-model transferability. TDAE adopts a bimodal optimization strategy that synergistically enhances image immunity by concurrently optimizing both visual and textual components. This innovative approach is achieved through two core mechanisms: the FlatGrad Defense Mechanism (FDM) at the image level and the Dynamic Prompt Defense (DPD) at the text level. The overall architecture of this process is visualized in Fig.~\ref{fig:framework}, while the step-by-step procedure is detailed in Algorithm~\ref{alg:tdae}. Together, these components enable TDAE to generate immunized samples that exhibit robust and generalizable protection against malicious image edits across a wide range of unseen editing models.
\begin{algorithm}[t!]
\caption{Transferable Defense Against Malicious Image Edits (TDAE)} 
\label{alg:tdae}
\KwIn{Clean image $x_0$, original prompt embedding $c$, loss function $\mathcal{L}$, image perturbation budget $\epsilon_v$, image step size $\alpha$, total iterations $N$, FDM regularization coefficient $\lambda$, DPD period $S$, text perturbation budget $\epsilon_p$, text step size $\eta$, DPD iterations $M$, benign target output $y_0$, FDM gradient step $h$}
\KwOut{Immunized image $x_{\text{adv}}$}
Initialize image perturbation $\delta_v \leftarrow \mathbf{0}$\;
Initialize text perturbation $\delta_p \leftarrow \mathbf{0}$\;
Initialize current immunized image $x_{\text{imu}} \leftarrow x_0$\; 

\For{$n = 1$ \KwTo $N$}{
    Set current text embedding for this iteration: $e \leftarrow c + \delta_p$\; 
    
    \If{$n \equiv 0 \pmod{S}$}{ 
        $\delta_p \leftarrow \mathbf{0}$\; 
        \For{$m = 1$ \KwTo $M$}{
            Let current text embedding for DPD be $e_{DPD} \leftarrow c + \delta_p$\;
            Compute DPD loss: $\mathcal{L}_{DPD} \leftarrow \mathcal{L}(f_\theta(x_{\text{imu}}, e_{DPD}), y_0)$\;
            Compute DPD gradient: $g_p \leftarrow \nabla_{\delta_p} \mathcal{L}_{DPD}$\;
            Update text perturbation: $\delta_p \leftarrow \delta_p - \eta \cdot \text{sign}(g_p)$\;
            Project text perturbation: $\delta_p \leftarrow \Pi_{\|\cdot\|_\infty \leq \epsilon_p}(\delta_p)$\;
        }
        $e \leftarrow c + \delta_p$\; 
    }
    
    Let $x_{\text{imu\_current}} \leftarrow x_0 + \delta_v$\;
    Compute base gradient: $g_1 \leftarrow \nabla_{\delta_v} \mathcal{L}(f_\theta(x_{\text{imu\_current}}, e), y_0)$\;
    Normalize direction: $\mathbf{s} \leftarrow g_1 / \|g_1\|_2$\;
    Let $\delta_v' \leftarrow \delta_v + h \mathbf{s}$\; 
    Let $x_{\text{imu}'} \leftarrow x_0 + \delta_v'$\;
    Compute gradient at perturbed point: $g_2 \leftarrow \nabla_{\delta_v} \mathcal{L}(f_\theta(x_{\text{imu}'}, e), y_0) \Big|_{\delta_v = \delta_v'}$\;
    Compute loss difference: $z \leftarrow \mathcal{L}(f_\theta(x_{\text{imu}'}, e), y_0) - \mathcal{L}(f_\theta(x_{\text{imu\_current}}, e), y_0)$\;
    
    Compute FDM gradient: $g_{FDM} \leftarrow -g_1 + \frac{\lambda}{h} \cdot \text{sign}(z) \cdot (g_2 - g_1)$\;
    
    Update image perturbation: $\delta_v \leftarrow \delta_v - \alpha \cdot \text{sign}(g_{FDM})$\;
    Project image perturbation: $\delta_v \leftarrow \Pi_{\|\cdot\|_\infty \leq \epsilon_v}(\delta_v)$\;
    $x_{\text{imu}} \leftarrow x_0 + \delta_v$\; 
}
$x_{\text{adv}} \leftarrow x_0 + \delta_v$\;
\Return $x_{\text{adv}}$
\end{algorithm}

\subsection{FlatGrad Defense Mechanism}
\label{fdm}
As discussed in Section~\ref{trans}, the TPA method~\cite{Fan2023TransferabilityBT} enhances the transferability of immunized images by minimizing the expected gradient norm within a neighborhood of the perturbed image. This process is formulated as:
\begin{equation}
\begin{aligned}
    \label{eq:tpa_objective}
    \min_{\|\delta_v\|_p \leq \epsilon_v} -\mathcal{L}&(f_\theta(x_0 + \delta_v, c), y_0) + \\&\lambda \mathbb{E}_{\Delta \sim \mathcal{N}(x_0+\delta_v)}  \left[ \|\nabla \mathcal{L}(f_\theta(\Delta, c), y_0)\|_2 \right],
\end{aligned}
\end{equation}
Here, $\lambda$ is a regularization hyperparameter, $c$ is the original text embedding, and $y_0$ is the benign target output. The variable $\Delta$ represents an image sampled from a local neighborhood $\mathcal{N}(x_0+\delta_v)$ centered at the perturbed image $x_0 + \delta_v$.

The optimization in Eq.~\eqref{eq:tpa_objective} guides immunized samples towards flat minima in the loss landscape, and its effectiveness has been empirically validated. However, its reliance on sampling-based estimation of the expected gradient norm incurs significant computational overhead. To address this limitation, FDM replaces this expectation minimization with the direct optimization of the \textit{maximum} gradient norm within a small local region around the current perturbation. This leads to the following objective for FDM.
\begin{equation}
\label{eq:fdm_obj1}
\begin{aligned}
    \min_{\|\delta_v\|_p \leq \epsilon_v} -\mathcal{L}&(f_\theta(x_{0} + \delta_v, e), y_0) +  \\ &\lambda \cdot \max_{\|\delta_v' - \delta_v\|_q \leq \rho}  \|\nabla_{\delta_v'} \mathcal{L}(f_\theta(x_{0} + \delta_v', e), y_0) \|_2,
\end{aligned}
\end{equation}
where \(e\) represents the current text embedding, which can be the original \(c\) or the DPD-refined \(c+\delta_p\), as detailed in Algorithm~\ref{alg:tdae}. \(\rho > 0\) defines the radius of a small local region under an \(L_q\) norm, often \(L_\infty\), around the current image perturbation \(\delta_v\). The first term aims to maximize the immunization effect by minimizing its negative. The second term regularizes the solution towards regions where the gradient norm is small, approaching a characteristic of flat minima. If the optimization converges to a flat minimum \(\delta_v^*\), then \(\nabla_{\delta_v} \mathcal{L}(f_\theta(x_0+\delta_v^*, e), y_0) \approx \mathbf{0}\). Consequently, the maximum gradient norm in its vicinity, \(\sup_{\|\delta_v' - \delta_v^*\|_q \leq \rho} \|\nabla_{\delta_v'} \mathcal{L}(f_\theta(x_{0} + \delta_v',e), y_0) \|_2\), will also be small, aligning with the objective in Eq.~\eqref{eq:fdm_obj1}.

Directly optimizing the inner maximization problem in Eq.~\eqref{eq:fdm_obj1} is computationally intractable as it would require solving a complex optimization subproblem at each step. To devise a practical and efficient algorithm while still promoting flatness, we propose a more \textit{tractable objective}. We simplify the regularization term by directly penalizing the gradient norm at the current point $\delta_v$, as shown in Eq.~\eqref{eq:fdm_obj2}. This simplification is motivated by the fact that a small gradient norm at $\delta_v$ is a \textit{necessary, albeit not sufficient}, condition for a small maximum gradient norm in its local vicinity. This modified objective still effectively guides the optimization towards flatter regions of the loss landscape.
\begin{equation}
\label{eq:fdm_obj2}
\begin{aligned}
     \min_{\|\delta_v\|_p \leq \epsilon_v}  -\mathcal{L}_{\delta_v} + \lambda \cdot \|\nabla_{\delta_v} \mathcal{L}_{\delta_v} \|_2,
\end{aligned}
\end{equation}
where we use the shorthand \(\mathcal{L}_{\delta_v} = \mathcal{L}(f_\theta(x_{0} + \delta_v, e), y_0)\) for notational brevity in subsequent derivations.

To optimize Eq.~\eqref{eq:fdm_obj2} using gradient-based methods, we need its gradient with respect to \(\delta_v\). Computing the gradient of the first term, $-\mathcal{L}_{\delta_v}$, yields $-\nabla_{\delta_v} \mathcal{L}_{\delta_v}$. For the second term, the regularization term \(\lambda \|\nabla_{\delta_v} \mathcal{L}_{\delta_v} \|_2\), we consider the fundamental definition of a gradient. The gradient magnitude \(\|\nabla_{\delta_v} \mathcal{L}_{\delta_v} \|_2\) represents the maximum rate of change of the loss function \(\mathcal{L}_{\delta_v}\). This rate of change occurs along the direction of the gradient itself. We can approximate this term by considering the change in loss along a specific direction. Drawing inspiration from methods that promote flatness by exploring the local loss landscape, such as SAM~\cite{Foret2020SharpnessAwareMF}, we choose the direction \(\mathbf{s}\) to be the normalized gradient direction. That is,
\begin{equation}
    \mathbf{s} = \frac{\nabla_{\delta_v} \mathcal{L}_{\delta_v}}{\|\nabla_{\delta_v} \mathcal{L}_{\delta_v}\|_2},
    \label{eq:s_direction_definition} 
\end{equation}
assuming \(\nabla_{\delta_v} \mathcal{L}_{\delta_v} \neq \mathbf{0}\); otherwise, \(\mathbf{s} = \mathbf{0}\).
With this chosen direction \(\mathbf{s}\), the \(L_2\) norm of the gradient can be expressed as the directional derivative along \(\mathbf{s}\), which is \(\mathbf{s}^T \nabla_{\delta_v} \mathcal{L}_{\delta_v} = \|\nabla_{\delta_v} \mathcal{L}_{\delta_v}\|_2\). This directional derivative can then be approximated using a finite difference for a small step size \(h > 0\):
\begin{equation}
    \|\nabla_{\delta_v} \mathcal{L}_{\delta_v}\|_2 \approx \frac{\mathcal{L}(f_\theta(x_0 + \delta_v + h\mathbf{s}, e), y_0) - \mathcal{L}_{\delta_v}}{h}.
    \label{eq:directional_derivative_approx}
\end{equation}

Let \(\mathcal{L}_{\delta_v+h\mathbf{s}} = \mathcal{L}(f_\theta(x_0 + \delta_v + h\mathbf{s}, e), y_0)\). Substituting the approximation from Eq.~\eqref{eq:directional_derivative_approx} into the regularization term \(\lambda \|\nabla_{\delta_v} \mathcal{L}_{\delta_v} \|_2\) of Eq.~\eqref{eq:fdm_obj2}, we get approximately \(\lambda \frac{\mathcal{L}_{\delta_v+h\mathbf{s}} - \mathcal{L}_{\delta_v}}{h}\). 
However, the norm \(\|\nabla_{\delta_v} \mathcal{L}_{\delta_v} \|_2\) is always non-negative. To ensure our approximation of this term also reflects this non-negativity and to provide robustness, particularly when \(h\) is small or due to numerical precision, we consider the absolute value of the difference. Thus, the objective function in Eq.~\eqref{eq:fdm_obj2} is approximated as:
\begin{equation}
    \min_{\|\delta_v\|_p \leq \epsilon_v}  -\mathcal{L}_{\delta_v} + \frac{\lambda}{h} \left| \mathcal{L}_{\delta_v+h\mathbf{s}} - \mathcal{L}_{\delta_v} \right|.
    \label{eq:fdm_obj_approx_abs} 
\end{equation}

We compute the gradient of this approximated objective with respect to \(\delta_v\). The gradient of the first term becomes \(-\nabla_{\delta_v} \mathcal{L}_{\delta_v}\). For the second term, using the chain rule \(\nabla_x |u| = \text{sign}(u) \nabla_x u\), and letting \(z = \mathcal{L}_{\delta_v+h\mathbf{s}} - \mathcal{L}_{\delta_v}\), the gradient of \(\frac{\lambda}{h} |z|\) is \(\frac{\lambda}{h} \text{sign}(z) (\nabla_{\delta_v}\mathcal{L}_{\delta_v+h\mathbf{s}} - \nabla_{\delta_v}\mathcal{L}_{\delta_v})\).
Therefore, the total gradient \(g_{FDM}\) of the objective in Eq.~\eqref{eq:fdm_obj_approx_abs} with respect to \(\delta_v\) is:
\begin{equation}
    \label{eq:fdm_final_gradient_form} 
    g_{FDM} = -\nabla_{\delta_v} \mathcal{L}_{\delta_v} +  \frac{\lambda}{h}\text{sign}(z) \left(\nabla_{\delta_v}\mathcal{L}_{\delta_v+h\mathbf{s}} - \nabla_{\delta_v}\mathcal{L}_{\delta_v}\right).
\end{equation}

The term \(\nabla_{\delta_v}\mathcal{L}_{\delta_v+h\mathbf{s}}\) involves the gradient at the perturbed point \(x_0 + \delta_v + h\mathbf{s}\). Following~\cite{Foret2020SharpnessAwareMF}, this can be estimated efficiently without computing second-order derivatives. First, \(\mathbf{s}\) is computed and we set \(\delta_v' = \delta_v + h\mathbf{s}\). The required gradient is then computed as follows:
\begin{equation}
\label{eq:fdm_hessian_free_approx}
\nabla_{\delta_v}\mathcal{L}_{\delta_v+h\mathbf{s}} \approx \left. \nabla_{\tilde{\delta}_v} \mathcal{L}(f_\theta(x_0 + \tilde{\delta}_v, e), y_0) \right|_{\tilde{\delta}_v = \delta_v'}
\end{equation}

This derived gradient \(g_{FDM}\) allows for an efficient, Hessian-free optimization of our flatness-promoting objective. It is then incorporated into a standard Projected Gradient Descent update for \(\delta_v\), as detailed in Algorithm~\ref{alg:tdae}.

\subsection{Dynamic Prompt Defense}
The FlatGrad Defense Mechanism (FDM) enhances the transferability of immunized samples at the image level by steering perturbations toward flat minima. However, its operation considers a fixed text embedding \(e\) during each FDM optimization step. This poses a potential vulnerability, as different editing models naturally exhibit subtle but critical variations in their understanding of the same text prompt. An image immunized against only a single, fixed interpretation of a prompt is therefore likely to be brittle and fail against unseen models. To address this, we propose Dynamic Prompt Defense (DPD), a mechanism designed to compel the image perturbation to become robust against a neighborhood of semantic interpretations, not just a single point.

Essentially, DPD introduces a dynamic adversarial process that explores the text embedding space to forge a more universally effective visual defense. It achieves this through a principled, cyclical adversarial process. First, with the image perturbation fixed, we dynamically update the text embedding. This forces the attacker model to find a new textual interpretation that can bypass the current immune features. Second, this newly discovered text embedding is fixed, and the image perturbation is updated to enhance immunity against malicious edits. This compels the visual attack to adapt and become stronger against both the original prompt and its semantic variant. By iterating this process, the image perturbation learns a broader set of immunity-enhancing features, significantly improving its generalization to diverse and unseen editing models.

Specifically, DPD acts as a complementary mechanism to FDM within the TDAE framework. Conventional approaches typically optimize the visual perturbation $\delta_v$ under a static text embedding. In contrast, our DPD introduces a learnable textual perturbation $\delta_p$ to the original text embedding $c$, forming an augmented embedding $c + \delta_p$. Periodically, at intervals of \(S\) FDM optimization iterations as shown in Algorithm~\ref{alg:tdae}, the image perturbation \(\delta_v\) is temporarily fixed. During this phase, DPD optimizes \(\delta_p\) with the objective of minimizing the immunization loss. That is, it seeks a perturbation \(\delta_p\) targeting the editing process. This perturbation forces the outcome \(f_\theta(x_0 + \delta_v, c + \delta_p)\) to align as closely as possible with the benign reference output \(y_0\). In other words, this step searches for a modified text prompt to bypass the current defense, thereby temporarily making the malicious edit work again.

Subsequently, this DPD-optimized text embedding \(e = c + \delta_p\) is used in the following FDM iterations to update the image perturbation \(\delta_v\). This process facilitates the periodic adaptation of \(\delta_v\) to an evolving set of challenging text embeddings. Such an alternating optimization allows for the systematic extraction of broader and more diversified immunity-relevant features by synchronizing \(\delta_v\) with progressively refined textual semantics. This, in turn, enhances the transferability and overall immunization effectiveness of the generated samples against unseen models.

Formally, let \(c\) denote the original text embedding extracted from a prompt, and let \(\delta_p\) represent the perturbation applied to \(c\). The magnitude of \(\delta_p\) is constrained within an \(\epsilon_p\)-neighborhood, usually under the \(L_\infty\)-norm \(\|\delta_p\|_\infty \leq \epsilon_p\), to preserve semantic integrity. The DPD optimization alternates between two main objectives.

First, for Text Embedding Optimization, \(\delta_p\) is adjusted to minimize the discrepancy between the editing result of the currently immunized image and the benign output, while keeping \(\delta_v\) unchanged. This corresponds to:
\begin{equation}
\label{eq:dpd_obj1}
\min_{\|\delta_p\|_\infty \leq \epsilon_p} \mathcal{L}\left( f_\theta(x_0 + \delta_v, c + \delta_p), y_0 \right),
\end{equation}
where \(\mathcal{L}\) is the loss function used throughout the framework. This step is iteratively solved, for instance, using Projected Gradient Descent as indicated in Algorithm~\ref{alg:tdae}.

Second, for Image Perturbation Optimization under the refined text embedding, the FDM objective Eq.~\eqref{eq:fdm_obj2} is then minimized with respect to \(\delta_v\) using the DPD-optimized text embedding \(e = c + \delta_p\), which is formulated as follows:
\begin{equation}
\label{eq:dpd_fdm_obj}
\begin{aligned} 
\min_{\|\delta_v\|_p \leq \epsilon_v} -\mathcal{L} &(f_\theta(x_{0} + \delta_v, c + \delta_p), y_0) \\
&+ \lambda \cdot \|\nabla_{\delta_v} \mathcal{L}(f_\theta(x_{0} + \delta_v, c + \delta_p), y_0) \|_2 
\end{aligned}
\end{equation}

This iterative, synergistic optimization between FDM and DPD allows TDAE to explore a richer solution space for \(\delta_v\), leading to enhanced robustness and transferability.

\section{Experiments}
\subsection{Experimental Setup}
\paragraph{Threat Model}In this paragraph, we define the capabilities and goals for both the defenders and the attackers as follows:

Attacker (Malicious User). The attacker obtains the publicly available immunized images and attempts to apply malicious edits in a black-box setting. They have no knowledge of the original image, the perturbation, the immunization method, or the surrogate model used for image immunization, but have full control over their chosen editing model, which may differ from the defender's surrogate. Our experiments simulate the attacker using standard inference settings (e.g., default schedulers, CFG scale, step counts from the original editing models) to achieve promising edit quality.

Defender (Image Owner). The defender has white-box access to a surrogate diffusion-based editing model to generate an imperceptible adversarial perturbation, aiming to immunize the image before public distribution. Our intra-model evaluations simulate the case where the defender uses the same editing model as the attacker, while our cross-model evaluations represent the more general black-box scenario where the defender uses the different editing model as the attacker.

\paragraph{Target Models and Dataset}
To evaluate our TDAE's efficacy, we conduct experiments on image immunization against leading open-source diffusion-based image editing models. Specifically, we test against multiple versions of StableDiffusion (SD) configured for editing tasks, including SD14~\cite{Rombach2021HighResolutionIS} and the recent SD3~\cite{Esser2024ScalingRF}, as well as the dedicated InstructPix2Pix (INS) model~\cite{Brooks2022InstructPix2PixLT}. These models are all accessed via the Hugging Face platform~\cite{huggingface}. Considering the absence of standardized benchmarks for evaluating image immunization techniques against such editing functionalities, we construct a dataset comprising 100 images meticulously selected from the publicly available \texttt{InstructPix2Pix-clip-filtered} collection~\cite{instructpix2pix-clip-filtered}, which are inherently suited for instruction-guided image editing. To ensure diversity and representativeness, this dataset is structured to include 35 portraits, 35 landscape images, and 30 artworks. For textual guidance during the editing process, we utilize the original editing instructions associated with each selected image within the source dataset, thereby maintaining the consistency and ensuring the relevance of the editing tasks to the inherent characteristics of the images and the models' editing capabilities.

\paragraph{Evaluation Settings and Metrics} 
To comprehensively assess the performance of TDAE, we design our experiments to cover both intra-model and cross-model evaluation scenarios. 

For intra-model evaluations, immunized images are generated by applying adversarial perturbations crafted against a specific source editing model. The effectiveness of these immunized images is then evaluated by attempting to edit them using the \textit{same} source model. These intra-model assessments are conducted on INS, SD3, and SD14, respectively. This setting evaluates the defense capability of an immunization method against the model it is directly optimized for.

For cross-model evaluations, we investigate the transferability of the immunized images generated by various methods. In this paradigm, adversarial perturbations are generated by surrogate models. The resulting immunized images are then subjected to editing attempts by one or more different target models that are unseen during the immunization generation process. This setup is crucial for gauging the generalization capability of a defense mechanism. Our cross-model evaluations are detailed in Tables~\ref{tab:my_comparison_table}, \ref{tab:sd3_comparison}, and \ref{tab:sd14_comparison}. These evaluations encompass several transfer scenarios, including immunization on INS with subsequent transfer to SD14 and SD3, immunization on SD3 followed by transfer to SD14 and INS, and immunization on SD14 with transfers to SD3 and INS. Such pairings are chosen to represent transfers between models exhibiting potentially significant architectural and training differences, thereby providing a rigorous test of cross-model robustness.

Across all evaluations, the immunization effectiveness is quantified using a suite of standard image quality and perceptual similarity metrics, which includes Peak Signal-to-Noise Ratio (PSNR), Structural Similarity Index Measure (SSIM)~\cite{1284395}, Visual Information Fidelity in Pixel domain (VIFP)~\cite{1576816}, and Feature Similarity Index Measure (FSIM)~\cite{5705575}. For these metrics, lower values indicate a more effective immunization, signifying greater distortion or deviation of the edited output from the benignly edited original image. Conversely, we also employ the Learned Perceptual Image Patch Similarity (LPIPS)~\cite{Zhang2018TheUE}, where higher values signify greater perceptual dissimilarity and thus more effective immunization. The arrows in the table headers denote the desired direction of change for each metric to indicate stronger immunization performance.

\subsection{Main Results}
To evaluate the effectiveness of our TDAE, we conduct extensive comparative experiments against several state-of-the-art image immunization methods. These baseline approaches include ACE~\cite{Zheng2023TargetedAI}, MIST~\cite{Liang2023MistTI}, PGD, PGE~\cite{Salman2023RaisingTC}, and SA~\cite{Lo_2024_CVPR}. Our evaluations are performed across three distinct and widely-used diffusion-based image editing models, namely InstructPix2Pix (INS), StableDiffusion v1-4 (SD14), and StableDiffusion v3 (SD3), covering both intra-model and cross-model evaluation. 

\begin{table*}[t!]
\centering
\caption{Quantitative comparison of immunization effectiveness when perturbations are generated on the InstructPix2Pix (INS) model. Results are shown for intra-model evaluation (INS) and cross-model transfer to StableDiffusion v1-4 (INS$\rightarrow$SD14) and StableDiffusion v3 (INS$\rightarrow$SD3). The arrows ($\uparrow$/$\downarrow$) indicate that higher/lower values of the metric signify better immunization performance.}
\label{tab:my_comparison_table} 
\resizebox{\textwidth}{!}{%
\begin{tabular}{l ccccc ccccc ccccc}
\toprule

& \multicolumn{5}{c}{\textbf{INS}} & \multicolumn{5}{c}{\textbf{INS$\rightarrow$SD14}} & \multicolumn{5}{c}{\textbf{INS$\rightarrow$SD3}} \\

\cmidrule(lr){2-6} \cmidrule(lr){7-11} \cmidrule(lr){12-16}

Method & PSNR $\downarrow$ & SSIM $\downarrow$ & LPIPS $\uparrow$ & VIFP $\downarrow$ & FSIM $\downarrow$ & PSNR $\downarrow$ & SSIM $\downarrow$ & LPIPS $\uparrow$ & VIFP $\downarrow$ & FSIM $\downarrow$ & PSNR $\downarrow$ & SSIM $\downarrow$ & LPIPS $\uparrow$ & VIFP $\downarrow$ & FSIM $\downarrow$ \\
\midrule

ACE~\cite{Zheng2023TargetedAI} & 16.12 & 0.5993 & 0.3686 & 0.2026 & 0.7872 & 16.05 & 0.3781 & 0.4156 & 0.0891 & 0.6849 & 20.94 & 0.6983 & 0.3362 & 0.2466 & 0.8482 \\
ACE~\cite{Zheng2023TargetedAI} + TDAE & 15.24 & 0.5828 & 0.4161 & 0.1840 & 0.7681 & 15.01 & 0.3421 & 0.4418 & 0.0692 & 0.6684 & 20.13 & 0.6745 & 0.3474 & 0.2301 & 0.8407 \\
MIST~\cite{Liang2023MistTI} & 16.51 & 0.6004 & 0.3675 & 0.2166 & 0.7927 & 16.40 & 0.4289 & 0.4148 & 0.0861 & 0.6927 & 21.71 & 0.7339 & 0.3120 & 0.2592 & 0.8577 \\
MIST~\cite{Liang2023MistTI} + TDAE & 15.39 & 0.5847 & 0.3961 & 0.2039 & 0.7760 & 15.53 & 0.4025 & 0.4301 & 0.0783 & 0.6824 & 20.83 & 0.7195 & 0.3367 & 0.2488 & 0.8436 \\
PGE~\cite{Salman2023RaisingTC} & 16.15 & 0.5945 & 0.3801 & 0.1858 & 0.7849 & 16.84 & 0.4153 & 0.4369 & 0.0899 & 0.7234 & 20.76 & 0.6578 & 0.3222 & 0.2467 & 0.8407 \\
PGE~\cite{Salman2023RaisingTC} + FDM & 15.49 & 0.5743 & 0.3982 & 0.1764 & 0.7715 & 16.07 & 0.3941 & 0.4497 & 0.0727 & 0.7065 & 20.06 & 0.6476 & 0.3325 & 0.2228 & 0.8337 \\
PGD~\cite{Salman2023RaisingTC} & 17.84 & 0.6658 & 0.3157 & 0.2408 & 0.8233 & 15.76 & 0.3781 & 0.4658 & 0.0762 & 0.7029 & 22.08 & 0.7276 & 0.2453 & 0.3168 & 0.8690 \\
PGD~\cite{Salman2023RaisingTC} + TDAE & 16.79 & 0.6038 & 0.3801 & 0.2077 & 0.8146 & 14.92 & 0.3517 & 0.4930 & 0.0672 & 0.6896 & 20.96 & 0.7008 & 0.2748 & 0.2914 & 0.8513 \\
SA~\cite{Lo_2024_CVPR} & 16.57 & 0.5982 & 0.3966 & 0.1913 & 0.7985 & 15.69 & 0.3570 & 0.4823 & 0.0731 & 0.6983 & 20.98 & 0.6516 & 0.3492 & 0.2419 & 0.8438 \\
SA~\cite{Lo_2024_CVPR} + TDAE & 15.32 & 0.5711 & 0.4396 & 0.1789 & 0.7794 & 15.04 & 0.3482 & 0.4893 & 0.0647 & 0.6880 & 19.64 & 0.6326 & 0.3664 & 0.2178 & 0.8310 \\
\bottomrule
\end{tabular}%
}
\end{table*}
\begin{figure*}[t]
    \centering
    \includegraphics[width=1\linewidth]{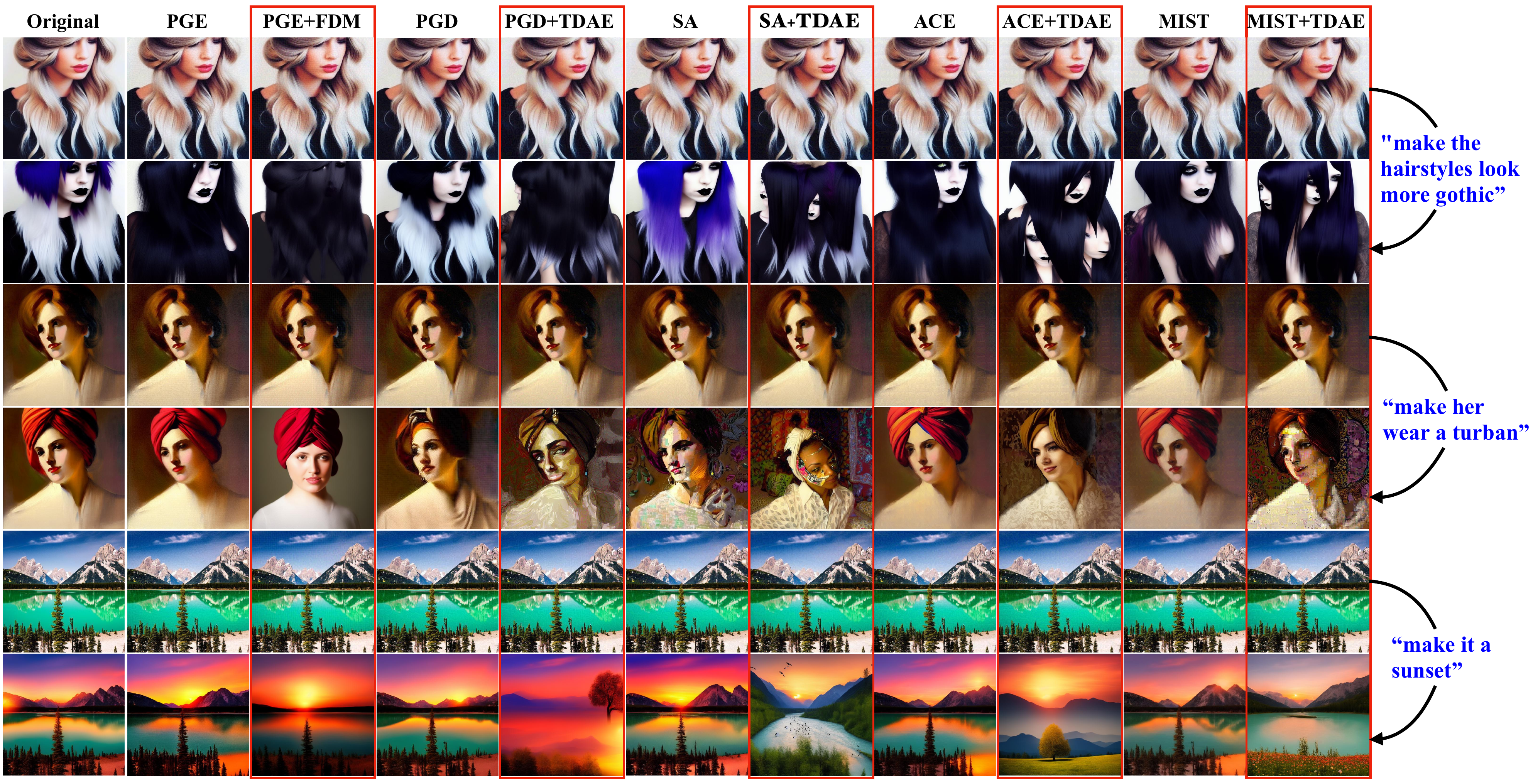}
    \caption{Visual comparison of editing outcomes on immunized images under an intra-model setting. The first column shows original images. Subsequent columns compare baseline immunization methods (PGE, PGD, SA, ACE, MIST) against their TDAE-enhanced (or FDM-enhanced for PGE) versions under various editing prompts. Enhanced methods typically lead to more pronounced editing failures or artifacts, indicating successful immunization.}
    \label{fig:intra}
\end{figure*}

\paragraph{Intra-Model Evaluation}

The immunization efficacy of our TDAE is first verified under intra-model settings. In this scenario, both the perturbation generation and evaluation are performed on the same editing model. Tables~\ref{tab:my_comparison_table} (INS), \ref{tab:sd3_comparison} (SD3), and \ref{tab:sd14_comparison} (SD14) demonstrate through intra-model evaluations that TDAE significantly enhances the protection strength of state-of-the-art immunization methods against their target models.

For INS evaluation (Table~\ref{tab:my_comparison_table}), TDAE integration with SA (SA + TDAE) shows substantial gains across all metrics: LPIPS increases from 0.3966 to 0.4396 while PSNR decreases from 16.57 to 15.32. Similar improvements occur with ACE + TDAE, MIST + TDAE, and PGD + TDAE compared to their baselines. Even PGE, which perturbs only the VAE encoder without text embedding interaction, achieves enhanced performance when combined with our FDM, increasing LPIPS from 0.3801 to 0.3982. Collectively, these results demonstrate consistent performance gains when augmenting baseline methods with our TDAE.

\begin{table*}[t!]
\centering
\caption{Quantitative comparison of immunization effectiveness when perturbations are generated on the StableDiffusion v3 (SD3) model. Results are shown for intra-model evaluation (SD3) and cross-model transfer to StableDiffusion v1-4 (SD3$\rightarrow$SD14) and InstructPix2Pix (SD3$\rightarrow$INS).}
\label{tab:sd3_comparison} 
\resizebox{\textwidth}{!}{%
\begin{tabular}{l ccccc ccccc ccccc}
\toprule

& \multicolumn{5}{c}{\textbf{SD3}} & \multicolumn{5}{c}{\textbf{SD3$\rightarrow$SD14}} & \multicolumn{5}{c}{\textbf{SD3$\rightarrow$INS}} \\

\cmidrule(lr){2-6} \cmidrule(lr){7-11} \cmidrule(lr){12-16}

Method & PSNR $\downarrow$ & SSIM $\downarrow$ & LPIPS $\uparrow$ & VIFP $\downarrow$ & FSIM $\downarrow$ & PSNR $\downarrow$ & SSIM $\downarrow$ & LPIPS $\uparrow$ & VIFP $\downarrow$ & FSIM $\downarrow$ & PSNR $\downarrow$ & SSIM $\downarrow$ & LPIPS $\uparrow$ & VIFP $\downarrow$ & FSIM $\downarrow$ \\
\midrule

PGE~\cite{Salman2023RaisingTC} & 18.14 & 0.5940 & 0.3796 & 0.2076 & 0.7921 & 16.63 & 0.4185 & 0.4434 & 0.0891 & 0.7246 & 18.50 & 0.6597 & 0.3378 & 0.2333 & 0.8311 \\
PGE~\cite{Salman2023RaisingTC} + FDM & 17.62 & 0.5861 & 0.3874 & 0.1919 & 0.7852 & 15.92 & 0.3959 & 0.4617 & 0.0819 & 0.7093 & 17.77 & 0.6324 & 0.3466 & 0.2261 & 0.8157 \\
PGD~\cite{Salman2023RaisingTC} & 18.43 & 0.5809 & 0.3921 & 0.2150 & 0.7890 & 16.25 & 0.3821 & 0.4622 & 0.0741 & 0.7045 & 18.61 & 0.6793 & 0.3013 & 0.2684 & 0.8304 \\
PGD~\cite{Salman2023RaisingTC} + TDAE & 17.38 & 0.5687 & 0.4109 & 0.1973 & 0.7721 & 15.54 & 0.3697 & 0.4806 & 0.0689 & 0.6903 & 17.33 & 0.6427 & 0.3395 & 0.2455 & 0.8224 \\
SA~\cite{Lo_2024_CVPR} & 17.85 & 0.5583 & 0.4225 & 0.1894 & 0.7643 & 15.75 & 0.3628 & 0.4760 & 0.0769 & 0.7004 & 18.60 & 0.6622 & 0.3457 & 0.2289 & 0.8359 \\
SA~\cite{Lo_2024_CVPR} + TDAE & 16.53 & 0.5224 & 0.4568 & 0.1631 & 0.7455 & 15.00 & 0.3413 & 0.4899 & 0.0706 & 0.6856 & 17.59 & 0.6317 & 0.3673 & 0.2020 & 0.8219 \\
\bottomrule
\end{tabular}%
}
\end{table*}

\begin{figure*}[t!]
    \centering
    \includegraphics[width=1\linewidth]{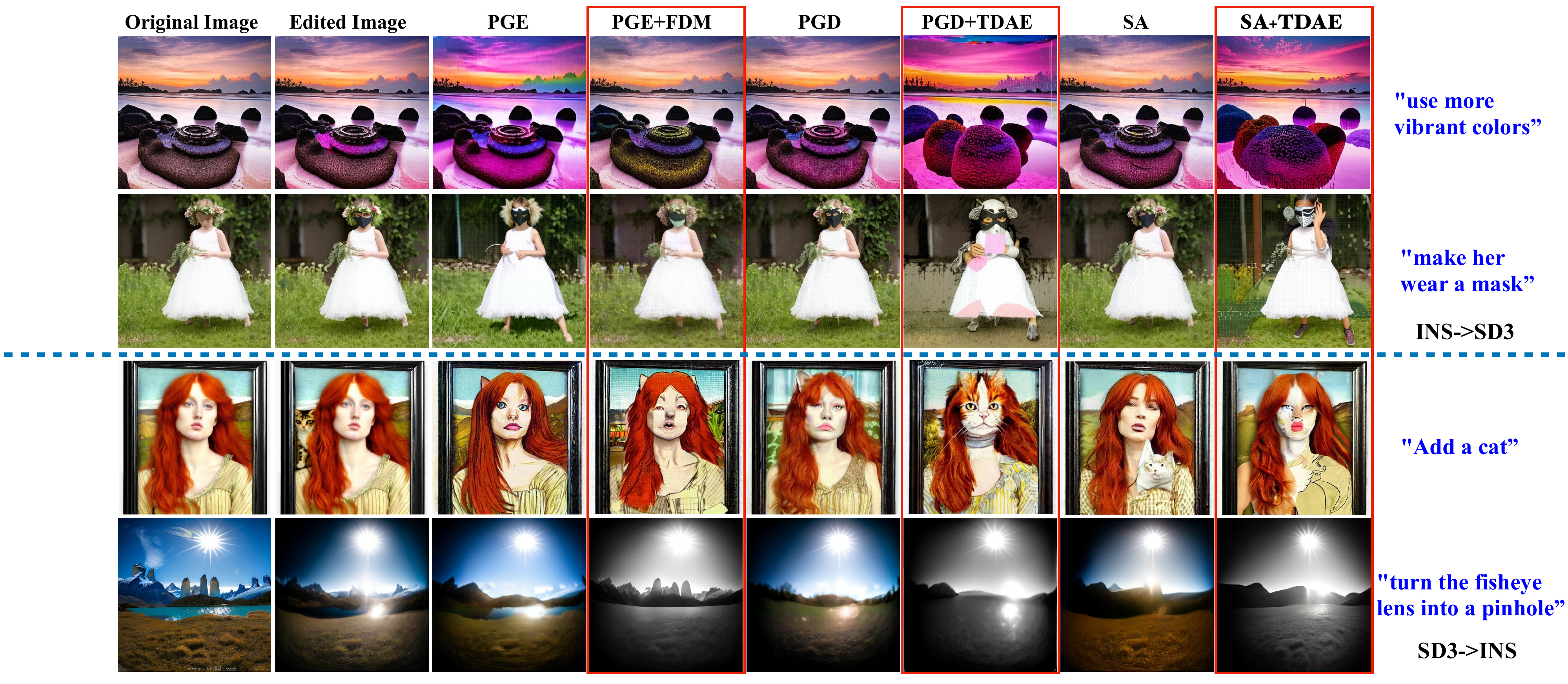}
    \caption{Qualitative assessment of cross-model transferability for TDAE-enhanced image immunization. The top two rows show results for the INS\(\rightarrow\)SD3 transfer (immunization on InstructPix2Pix, editing on StableDiffusion v3), while the bottom two rows depict the SD3\(\rightarrow\)INS transfer. Original and benignly edited images are provided for reference.}
    \label{fig:crosss}
\end{figure*} 

For SD3 evaluation (Table~\ref{tab:sd3_comparison}), TDAE-enhanced methods maintain superior immunization performance. SA + TDAE achieves the highest LPIPS of 0.456, while PGD + TDAE shows significant gains from 0.3921 to 0.4109. Similarly, PGE + FDM improves LPIPS (0.3874 vs. 0.3796). Note that ACE and MIST baselines are excluded not due to TDAE limitations, but because their architectures are inherently incompatible with SD3's Diffusion Transformer (DiT) design.

Consistent findings are observed with SD14 evaluations, presented in Table~\ref{tab:sd14_comparison}. PGD + TDAE achieves a substantially higher LPIPS (0.5504 vs. 0.4989), while SA + TDAE elevates LPIPS from 0.5285 to 0.5493. Across all baselines, TDAE/FDM integration enhances each metric, demonstrating stronger disruption of editing processes and improved immunization efficacy.

These quantitative improvements are further supported by qualitative comparisons, as illustrated in Fig.~\ref{fig:intra}. The qualitative results reveal that TDAE-enhanced immunizations yield significantly more pronounced and semantically disruptive artifacts during editing compared to baseline methods. For instance, when generating a ``gothic hairstyle'' (second row), TDAE-enhanced methods like SA + TDAE often produce severely distorted dark hair with unnatural coloration and textural loss, indicating a stronger defense. Similarly, for the ``make her wear a turban'' edit (fourth row), baseline methods typically yield results closely resembling benignly edited original images, exhibiting only minor subject deformations. In stark contrast, images immunized by TDAE-enhanced methods, subjected to the same edit, display either a drastically altered main subject or pervasive unnatural distortions, effectively neutralizing the edit. In landscape modifications, such as the ``make it a sunset'' (bottom two rows), TDAE-enhanced immunizations typically result in scenes with aberrant color palettes, object fragmentation, or a complete failure to depict a convincing sunset. A notable example is the SA + TDAE outcome, which transforms a lakeside sunset with distant mountains into an early morning valley scene, a fundamental alteration of both time and geography. This contrasts sharply with baseline methods that tend to preserve core structures while primarily modifying color schemes. Such visual evidence underscores TDAE's superior ability to degrade the quality and semantic coherence of malicious edits, consistent with the quantitative metrics.

\paragraph{Cross-Model Evaluation}
Robust image immunization critically depends on generalization to unseen editing models. We therefore evaluate the cross-model transferability of TDAE-enhanced immunizations. Adversarial perturbations are generated against a specific source model, and the immunized images are then edited by unseen target models. The quantitative results for these cross-model scenarios are detailed in Tables~\ref{tab:my_comparison_table}, \ref{tab:sd3_comparison}, and \ref{tab:sd14_comparison}. Across all evaluations, methods augmented with TDAE or its FDM component consistently demonstrate superior defensive performance compared to their baseline counterparts when transferred to unseen models.

For instance, when immunizing on INS and transferring to SD14, results are found in the INS$\rightarrow$SD14 columns of Table~\ref{tab:my_comparison_table}. Here, PGD + TDAE exhibits a substantial improvement in LPIPS from 0.4658 to 0.4930. This is accompanied by a notable reduction in PSNR from 15.76 to 14.92 and SSIM from 0.3781 to 0.3517, indicating stronger perceptual and structural disruption. Similarly, in the SD3 to INS transfer, detailed in the SD3$\rightarrow$INS columns of Table~\ref{tab:sd3_comparison}, SA + TDAE boosts LPIPS from 0.3457 to 0.3673, an increase of over six percent. Simultaneously, PSNR decreases by more than a full point from 18.60 to 17.59, and VIFP reduces from 0.2289 to 0.2020.

\begin{table*}[t!]
\centering
\caption{Quantitative comparison of immunization effectiveness when perturbations are generated on the StableDiffusion v1-4 (SD14) model. Results are shown for intra-model evaluation (SD14) and cross-model transfer to StableDiffusion v3 (SD14$\rightarrow$SD3) and InstructPix2Pix (SD14$\rightarrow$INS).}
\label{tab:sd14_comparison} 
\resizebox{\textwidth}{!}{%
\begin{tabular}{l ccccc ccccc ccccc}
\toprule

& \multicolumn{5}{c}{\textbf{SD14}} & \multicolumn{5}{c}{\textbf{SD14$\rightarrow$SD3}} & \multicolumn{5}{c}{\textbf{SD14$\rightarrow$INS}} \\

\cmidrule(lr){2-6} \cmidrule(lr){7-11} \cmidrule(lr){12-16}

Method & PSNR $\downarrow$ & SSIM $\downarrow$ & LPIPS $\uparrow$ & VIFP $\downarrow$ & FSIM $\downarrow$ & PSNR $\downarrow$ & SSIM $\downarrow$ & LPIPS $\uparrow$ & VIFP $\downarrow$ & FSIM $\downarrow$ & PSNR $\downarrow$ & SSIM $\downarrow$ & LPIPS $\uparrow$ & VIFP $\downarrow$ & FSIM $\downarrow$ \\
\midrule

ACE~\cite{Zheng2023TargetedAI} & 15.48 & 0.3403 & 0.4483 & 0.0730 & 0.6722 & 21.24 & 0.6305 & 0.3554 & 0.2365 & 0.8428 & 16.68 & 0.6270 & 0.3649 & 0.2226 & 0.8366 \\
ACE~\cite{Zheng2023TargetedAI} + TDAE & 14.57 & 0.2813 & 0.4861 & 0.0532 & 0.6582 & 20.27 & 0.6118 & 0.3692 & 0.2197 & 0.8293 & 15.92 & 0.6045 & 0.3907 & 0.2000 & 0.8177 \\
MIST~\cite{Liang2023MistTI} & 16.14 & 0.3994 & 0.4311 & 0.0795 & 0.6829 & 21.98 & 0.6473 & 0.3345 & 0.2449 & 0.8523 & 17.04 & 0.6337 & 0.3283 & 0.2494 & 0.7982 \\
MIST~\cite{Liang2023MistTI} + TDAE & 15.46 & 0.3698 & 0.4682 & 0.0607 & 0.6657 & 21.31 & 0.6394 & 0.3470 & 0.2267 & 0.8455 & 16.30 & 0.6196 & 0.3331 & 0.2303 & 0.7907 \\
PGE~\cite{Salman2023RaisingTC} & 16.02 & 0.3910 & 0.4658 & 0.0818 & 0.7126 & 20.72 & 0.6520 & 0.3258 & 0.2392 & 0.8377 & 16.83 & 0.6284 & 0.3647 & 0.1982 & 0.8055 \\
PGE~\cite{Salman2023RaisingTC} + FDM & 15.24 & 0.3537 & 0.4914 & 0.0617 & 0.6973 & 19.94 & 0.6287 & 0.3401 & 0.2266 & 0.8215 & 16.20 & 0.5927 & 0.3821 & 0.1896 & 0.7914 \\
PGD~\cite{Salman2023RaisingTC} & 15.34 & 0.3170 & 0.4989 & 0.0690 & 0.6760 & 21.22 & 0.6636 & 0.3318 & 0.2564 & 0.8440 & 18.33 & 0.6795 & 0.3037 & 0.2693 & 0.8317 \\
PGD~\cite{Salman2023RaisingTC} + TDAE & 14.70 & 0.2594 & 0.5504 & 0.0579 & 0.6541 & 20.04 & 0.6467 & 0.3483 & 0.2412 & 0.8370 & 17.67 & 0.6506 & 0.3537 & 0.2331 & 0.8237 \\
SA~\cite{Lo_2024_CVPR} & 15.21 & 0.2684 & 0.5285 & 0.0586 & 0.6551 & 20.90 & 0.6314 & 0.3634 & 0.2339 & 0.8340 & 17.69 & 0.6182 & 0.3927 & 0.2015 & 0.8171 \\
SA~\cite{Lo_2024_CVPR} + TDAE & 14.72 & 0.2556 & 0.5493 & 0.0505 & 0.6476 & 19.86 & 0.6199 & 0.3789 & 0.2231 & 0.8208 & 16.86 & 0.6011 & 0.4056 & 0.1936 & 0.8026 \\
\bottomrule
\end{tabular}%
}
\end{table*}

\begin{table*}[t!]
\centering
\caption{Comparison of FDM with TPA~\cite{Fan2023TransferabilityBT} in terms of immunization effectiveness and optimization time per iteration. Evaluations are conducted under intra-model (INS) and cross-model (INS$\rightarrow$SD3) settings when integrated with SA and PGD baselines.}
\label{tab:fdm_vs_tpa_comparison}
\resizebox{\textwidth}{!}{
\begin{tabular}{lcccc | cccc}
\toprule
& \multicolumn{4}{c}{\textbf{INS (Intra-Model)}} & \multicolumn{4}{c}{\textbf{INS$\rightarrow$SD3 (Cross-Model)}} \\
\cmidrule(lr){2-5} \cmidrule(lr){6-9}
Method & SA~\cite{Lo_2024_CVPR} + TPA & SA~\cite{Lo_2024_CVPR} + FDM & PGD~\cite{Salman2023RaisingTC} + TPA & PGD~\cite{Salman2023RaisingTC} + FDM & SA~\cite{Lo_2024_CVPR} + TPA & SA~\cite{Lo_2024_CVPR} + FDM & PGD~\cite{Salman2023RaisingTC} + TPA & PGD~\cite{Salman2023RaisingTC} + FDM \\
\midrule
PSNR $\downarrow$       & 15.98 & 16.02 & 17.40 & 17.33 & 19.87 & 20.03 & 21.36 & 21.67 \\
SSIM $\downarrow$       & 0.5834& 0.5870& 0.6511& 0.6472& 0.6389& 0.6412& 0.7092& 0.7103\\
LPIPS $\uparrow$      & 0.4136& 0.4115& 0.3478& 0.3429& 0.3594& 0.3556& 0.2546& 0.2524\\
\midrule
Opt. Time/Iter (s) $\downarrow$ & 71.37  & 6.49 & 89.65 & 8.83 & 71.37  & 6.49 & 89.65 & 8.83 \\
\bottomrule
\end{tabular}%
} 
\end{table*}

\begin{table*}[t!]
\centering
\caption{Imperceptibility assessment of immunized images on the INS model, comparing TDAE-enhanced (or FDM-enhanced for PGE) methods against baselines by measuring visual fidelity to original clean images. Higher PSNR, SSIM, VIFP, FSIM ($\uparrow$) and lower LPIPS ($\downarrow$) denote superior imperceptibility.}
\label{tab:invisibility_comparison_ins}
\resizebox{\textwidth}{!}{
\begin{tabular}{lcccccccccc}
\toprule
& \multicolumn{10}{c}{\textbf{INS (Comparison with Original Clean Image)}} \\
\cmidrule(lr){2-11}
Metric/Method & ACE~\cite{Zheng2023TargetedAI} & ACE~\cite{Zheng2023TargetedAI}+TDAE & MIST~\cite{Liang2023MistTI} & MIST~\cite{Liang2023MistTI}+TDAE & PGD~\cite{Salman2023RaisingTC} & PGD~\cite{Salman2023RaisingTC}+TDAE & PGE~\cite{Salman2023RaisingTC} & PGE~\cite{Salman2023RaisingTC}+TDAE & SA~\cite{Lo_2024_CVPR} & SA~\cite{Lo_2024_CVPR}+TDAE \\
\midrule
PSNR $\uparrow$  & 34.68 & 34.24 & 34.27 & 34.09 & 36.04 & 35.09 & 34.37 & 34.00 & 34.71 & 34.41 \\
SSIM $\uparrow$  & 0.9027& 0.8976& 0.8914& 0.8870& 0.9136& 0.8972& 0.8812& 0.8818& 0.8916& 0.8877\\
LPIPS $\downarrow$ & 0.2317& 0.2369& 0.2430& 0.2473& 0.2262& 0.2474& 0.2245& 0.2309& 0.2364& 0.2384\\
VIFP $\uparrow$  & 0.5782& 0.5674& 0.5685& 0.5630& 0.6230& 0.5919& 0.5527& 0.5456& 0.5648& 0.5629\\
FSIM $\uparrow$  & 0.9724& 0.9710& 0.9744& 0.9738& 0.9778& 0.9746& 0.9748& 0.9664& 0.9703& 0.9665\\
\bottomrule
\end{tabular}%
} 
\end{table*}
\begin{table*}[t!]
\centering
\caption{Ablation study on the contributions of FDM and DPD components within the TDAE framework, integrated with PGD and SA baselines. Evaluations are performed under intra-model (INS) and cross-model (INS$\rightarrow$SD3) settings. Lower PSNR, SSIM, VIFP, FSIM ($\downarrow$) and higher LPIPS ($\uparrow$) indicate stronger immunization.}
\label{tab:ablation_study_fdm_dpd}
\begin{tabular}{lcccc | cccc}
\toprule
& \multicolumn{4}{c}{\textbf{INS (Intra-Model)}} & \multicolumn{4}{c}{\textbf{INS$\rightarrow$SD3 (Cross-Model)}} \\
\cmidrule(lr){2-5} \cmidrule(lr){6-9}
Method & PGD~\cite{Salman2023RaisingTC} & PGD~\cite{Salman2023RaisingTC}+FDM & PGD~\cite{Salman2023RaisingTC}+DPD & PGD~\cite{Salman2023RaisingTC}+TDAE & PGD~\cite{Salman2023RaisingTC} & PGD~\cite{Salman2023RaisingTC}+FDM & PGD~\cite{Salman2023RaisingTC}+DPD & PGD~\cite{Salman2023RaisingTC}+TDAE \\
\midrule
PSNR  $\downarrow$ & 17.84 & 17.33 & 17.49 & 16.79 & 22.08 & 21.67 & 21.73 & 20.96 \\
SSIM  $\downarrow$ & 0.6658& 0.6472& 0.6318& 0.6038 & 0.7276& 0.7103& 0.7144& 0.7008\\
LPIPS $\uparrow$ & 0.3157& 0.3429& 0.3694& 0.3801 & 0.2453& 0.2524& 0.2618& 0.2748\\
VIFP $\downarrow$ & 0.2408& 0.2196& 0.2247& 0.2077 & 0.3168& 0.3072& 0.3061& 0.2914\\
FSIM $\downarrow$ & 0.8233& 0.8213& 0.8205& 0.8146 & 0.8690& 0.8614& 0.8592& 0.8513\\
\midrule
\midrule 
Method & SA~\cite{Lo_2024_CVPR} & SA~\cite{Lo_2024_CVPR}+FDM & SA~\cite{Lo_2024_CVPR}+DPD & SA~\cite{Lo_2024_CVPR}+TDAE & SA~\cite{Lo_2024_CVPR} & SA~\cite{Lo_2024_CVPR}+FDM & SA~\cite{Lo_2024_CVPR}+DPD & SA~\cite{Lo_2024_CVPR}+TDAE \\
\midrule
PSNR $\downarrow$ & 16.57 & 16.02 & 15.74 & 15.32 & 20.98 & 20.03 & 20.26 & 19.64 \\
SSIM $\downarrow$ & 0.5982& 0.5870& 0.5813& 0.5711 & 0.6516& 0.6412& 0.6459& 0.6326\\
LPIPS $\uparrow$ & 0.3966& 0.4115& 0.4278& 0.4396 & 0.3492& 0.3556& 0.3587& 0.3664\\
VIFP $\downarrow$ & 0.1913& 0.1874& 0.1836& 0.1789 & 0.2419& 0.2208& 0.2331& 0.2178\\
FSIM $\downarrow$ & 0.7985& 0.7891& 0.7842& 0.7794 & 0.8438& 0.8374& 0.8396& 0.8310\\
\bottomrule
\end{tabular}
\end{table*}

\begin{figure*}[t!]
    \centering
    \includegraphics[width=1\linewidth]{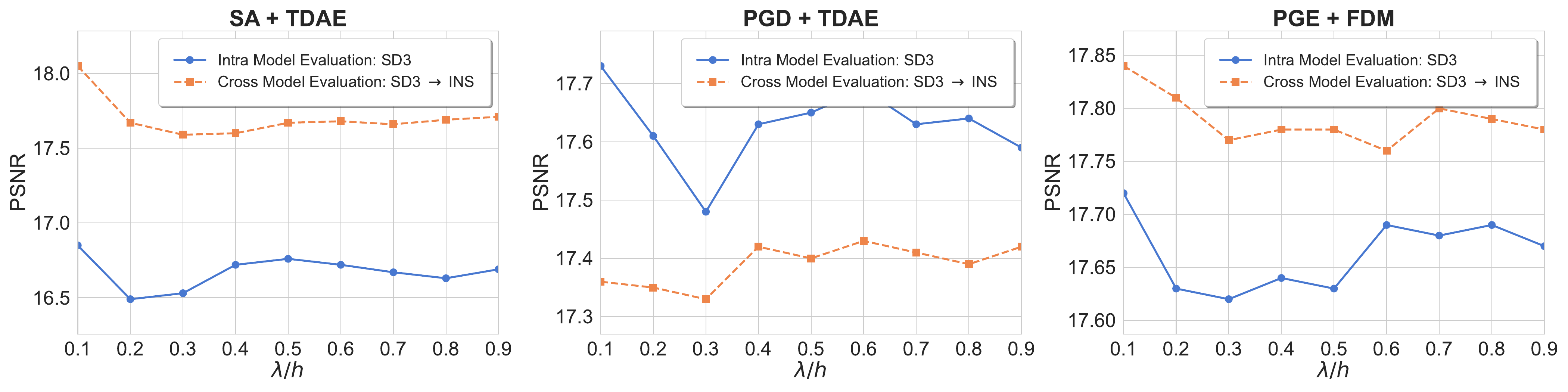}
    \caption{Ablation study on the hyperparameter \(\lambda/h\). This parameter is integral to TDAE when applied with PGD and SA, and to FDM when used with PGE. The plots illustrate the impact of varying \(\lambda/h\) on PSNR, where lower values signify better immunization effectiveness.}
    \label{fig:ablation_lambda_h}
\end{figure*} 

It is also pertinent to consider the inherent robustness of the target models. StableDiffusion v3 (SD3), with its advanced architecture, exhibits greater resilience to adversarial perturbations compared to earlier models like SD14 or INS. Consequently, immunized samples transferred to SD3 from other sources often show diminished defensive effects in quantitative metrics. For example, transferring perturbations crafted on SD14 to SD3 (Table~\ref{tab:sd14_comparison}, SD14 $\rightarrow$ SD3 columns) results in higher PSNR and lower LPIPS values than transfers to INS, even for baseline methods. Significantly, despite SD3's inherent robustness, TDAE integration still delivers substantial improvements. In this SD14 $\rightarrow$ SD3 transfer, SA+TDAE increases LPIPS from 0.3634 to 0.3789 and markedly reduces PSNR from 20.90 to 19.86, while FSIM decreases from 0.8340 to 0.8208. Furthermore, FDM, as a TDAE component, alone consistently enhances cross-model defense. For instance, PGE+FDM improves LPIPS from 0.3647 to 0.3821 in the SD14 $\rightarrow$ INS transfer. These results demonstrate TDAE's robust capability to generalize effectively across diverse model pairings and metrics, providing meaningful gains even against highly robust targets like SD3.

Qualitative comparisons in Fig.~\ref{fig:crosss} further substantiate TDAE's superior cross-model transferability. Baseline methods often yield edited images resembling benign edits of originals, while TDAE-enhanced methods consistently induce severe editing failures with substantial semantic distortions. For instance, during the INS to SD3 transfer with the ``use more vibrant colors'' prompt, TDAE-enhanced images significantly deviate from expectations or introduce unintended background alterations despite partial edit success. The PGE+FDM method notably degrades the image into abstract color fields. A similar pattern is observed in the SD3 to INS transfer for ``turn the fisheye lens into a pinhole''. While the immunized images generated by the baseline method exhibit slight distortions during editing, they generally achieve the desired ``turn the fisheye lens into a pinhole'' effect. In contrast, although the TDAE-enhanced method also successfully implements the ``turn the fisheye lens into a pinhole'' edit, the overall color palette of the image dramatically shifts from blue to a dark gray. This consistent induction of severe distortions, semantic incongruities or undesirable global changes demonstrates TDAE's robust capability to disrupt malicious edits on unseen models.

\subsection{Comparison with TPA and Efficiency Analysis}
\label{sec:comparison_tpa_efficiency}

To further validate the design choices within our FDM, particularly its efficiency and effectiveness in achieving flat minima, we conduct a comparative analysis against the state-of-the-art TPA method~\cite{Fan2023TransferabilityBT}. Both methods aim to find flat minima to enhance transferability. However, TPA relies on computationally intensive expectation minimization over neighborhood samples. FDM, by approximating this via local gradient norm optimization, is designed for greater efficiency.

Quantitative results are shown in Table~\ref{tab:fdm_vs_tpa_comparison}. When integrated with SA and PGD baselines under both intra-model (INS) and cross-model (INS$\rightarrow$SD3) settings, FDM achieves immunization performance comparable to TPA. Crucially, FDM exhibits a significant advantage in computational efficiency. As shown in Table~\ref{tab:fdm_vs_tpa_comparison}, FDM-integrated methods are considerably faster per iteration. SA + FDM requires only 6.49 seconds per iteration, whereas SA + TPA takes 71.37 seconds. Similarly, PGD+FDM completes iterations in 8.83 seconds versus PGD+TPA's 89.65 seconds. This substantial efficiency gain stems from FDM's elimination of TPA's computationally intensive multi-sample averaging. By matching TPA's defensive performance while reducing optimization time by an order of magnitude, FDM delivers highly efficient transferable immunization.

\subsection{Imperceptibility of Immunized Images}
\label{sec:imperceptibility}
Maintaining imperceptible adversarial perturbations is crucial for practical image immunization, ensuring visual fidelity is preserved. We quantitatively evaluate the visual similarity between the original clean images and their immunized counterparts generated by various baseline methods, both with and without TDAE or FDM. These comparisons are conducted on immunizations generated using the INS model, and the results are presented in Table~\ref{tab:invisibility_comparison_ins}. Higher values for PSNR, SSIM, VIFP, and FSIM, along with lower LPIPS values, indicate superior imperceptibility through closer resemblance to originals.

The results in Table~\ref{tab:invisibility_comparison_ins} consistently demonstrate that the integration of TDAE or FDM does not significantly degrade the visual quality of the immunized images compared to the baseline methods. For example, the PSNR value for SA is 34.71, and for SA + TDAE, it is 34.41. While this represents a very slight decrease, the absolute PSNR values remain high, generally above 34 dB for most TDAE-enhanced methods, which is typically considered indicative of high visual fidelity. Similarly, SSIM values for TDAE-enhanced methods, such as 0.8976 for ACE + TDAE and 0.8877 for SA + TDAE, remain very close to their respective baselines and are indicative of strong structural similarity.

The LPIPS metric, which measures perceptual similarity, also shows minimal changes. For instance, LPIPS for ACE is 0.2317, and for ACE + TDAE, it is 0.2369. Such small differences in LPIPS suggest that the perceptual impact of the additional perturbations introduced by TDAE is negligible. Similar trends are observed for VIFP and FSIM across all compared methods. For example, MIST achieves an FSIM of 0.9744, while MIST + TDAE records an FSIM of 0.9738. These consistently high similarity scores across multiple metrics affirm that the perturbations generated by TDAE-enhanced approaches, while effective in thwarting malicious edits, remain largely imperceptible. This ensures that the primary goal of protecting images without aesthetically compromising them is well achieved, making TDAE a practical solution for real-world applications.

\subsection{Ablation Study}
\label{sec:ablation_study} 
To rigorously evaluate the contributions of the core components within our TDAE framework and to understand the impact of key hyperparameters, we conduct a series of ablation studies. These studies specifically investigate the individual and combined roles of the FDM and the DPD, as well as the sensitivity of TDAE's performance to the hyperparameter \(\lambda/h\).

\paragraph{Effectiveness of FDM and DPD} 
\label{sec:ablation_components}
We first analyze the distinct contributions of FDM and DPD to the overall effectiveness of TDAE, particularly in enhancing cross-model transferability. These experiments involve integrating FDM alone, DPD alone, and both components synergistically (full TDAE) with strong baseline methods like SA and PGD. The quantitative results of this component ablation are presented in Table~\ref{tab:ablation_study_fdm_dpd}. 

The results presented in Table~\ref{tab:ablation_study_fdm_dpd} clearly illustrate the benefits of each component. When FDM is solely integrated with SA, a configuration denoted SA + FDM, we observe an improvement in immunization performance over the baseline SA. This improvement is evident in both intra-model INS and cross-model INS$\rightarrow$SD3 settings. For instance, in the INS$\rightarrow$SD3 transfer scenario, SA + FDM reduces PSNR to 20.03 from SA's 20.98 and also increases LPIPS. Similarly, integrating only DPD, as in the SA + DPD setup, enhances robustness. This is exemplified by an LPIPS of 0.4278 in the intra-model INS setting, compared to SA's 0.3966. However, the most significant gains are consistently achieved when both FDM and DPD are employed concurrently within the full TDAE framework. Taking SA + TDAE as an example, this complete configuration further reduces PSNR to 15.32 for intra-model evaluation and 19.64 for cross-model INS$\rightarrow$SD3 transfer. It also boosts LPIPS to 0.4396 and 0.3664 in these respective settings, outperforming configurations that utilize only a single component. This synergistic interaction demonstrates that FDM's flat minima seeking capabilities and DPD's diversified feature extraction approach complementarily address the challenge of transferable defense against malicious image edits.

\paragraph{Sensitivity to Hyperparameter \(\lambda/h\)}
\label{sec:ablation_lambda_h_hyperparameter} 
The hyperparameter \(\lambda/h\) within our FDM formulation (and thus TDAE) plays a critical role in balancing the primary immunization loss with the flatness regularization term. To determine an optimal operational range and understand its influence, we conduct an ablation study by varying the value of \(\lambda/h\). The results, illustrating the impact on PSNR for different TDAE-enhanced methods under both intra-model and cross-model evaluations, are presented in Fig.~\ref{fig:ablation_lambda_h}.

As depicted in Fig.~\ref{fig:ablation_lambda_h}, the PSNR values, where lower indicates better immunization, exhibit sensitivity to the choice of \(\lambda/h\). For methods like PGD + TDAE and SA + TDAE, both intra-model (SD3) and cross-model (SD3 to INS transfer) evaluations suggest that an optimal or near-optimal performance is achieved when \(\lambda/h\) is around 0.3. At this value, a favorable balance between robust intra-model defense and strong cross-model transferability is observed, as indicated by the lower PSNR values. Values of \(\lambda/h\) that are too small may not sufficiently enforce flatness, while excessively large values might over-penalize the primary immunization objective. The PGE + FDM configuration shows a slightly different sensitivity profile, but overall, these findings guide the selection of an effective \(\lambda/h\) for TDAE, confirming its impact on achieving robust defensive efficacy. Based on these results, we adopt \(\lambda/h = 0.3\) for other experiments involving TDAE with PGD and SA.

\section{Conclusion}
\label{sec:conclusion}
This paper introduces Transferable Defense Against Malicious Image Edits (TDAE), a novel bimodal framework that functions as a plug-and-play module to significantly enhance both the robustness and cross-model transferability of image immunizations against diffusion-based editing models. TDAE empowers existing immunization methods by synergistically integrating an image-level FlatGrad Defense Mechanism (FDM) for finding generalizable flat minima efficiently, and a text-level Dynamic Prompt Defense (DPD) for discovering broader immunity features. Extensive evaluations unequivocally demonstrate TDAE's superiority in intra-model and particularly cross-model scenarios, with augmented methods showing substantially stronger resistance to edits from unseen models while maintaining perturbation imperceptibility. Ablation studies validate the complementary contributions of FDM and DPD. This work offers a more practical and universally effective defense against the misuse of advanced image editing technologies, paving the way for future explorations into even more robust and efficient immunization strategies across diverse generative modalities.

\bibliographystyle{IEEEtran}

\bibliography{main}

\begin{thebibliography}{10}
\providecommand{\url}[1]{#1}
\csname url@samestyle\endcsname
\providecommand{\newblock}{\relax}
\providecommand{\bibinfo}[2]{#2}
\providecommand{\BIBentrySTDinterwordspacing}{\spaceskip=0pt\relax}
\providecommand{\BIBentryALTinterwordstretchfactor}{4}
\providecommand{\BIBentryALTinterwordspacing}{\spaceskip=\fontdimen2\font plus
\BIBentryALTinterwordstretchfactor\fontdimen3\font minus \fontdimen4\font\relax}
\providecommand{\BIBforeignlanguage}[2]{{%
\expandafter\ifx\csname l@#1\endcsname\relax
\typeout{** WARNING: IEEEtran.bst: No hyphenation pattern has been}%
\typeout{** loaded for the language `#1'. Using the pattern for}%
\typeout{** the default language instead.}%
\else
\language=\csname l@#1\endcsname
\fi
#2}}
\providecommand{\BIBdecl}{\relax}
\BIBdecl

\bibitem{Ho2020DenoisingDP}
J.~Ho, A.~Jain, and P.~Abbeel, ``{Denoising Diffusion Probabilistic Models},'' \emph{arXiv}, vol. abs/2006.11239, 2020.

\bibitem{SohlDickstein2015DeepUL}
J.~Sohl-Dickstein, E.~A. Weiss, N.~Maheswaranathan, and S.~Ganguli, ``{Deep Unsupervised Learning using Nonequilibrium Thermodynamics},'' in \emph{Proc. Int. Conf. Mach. Learn.}, 2015.

\bibitem{Song2020ScoreBasedGM}
Y.~Song, J.~Sohl-Dickstein, D.~P. Kingma, A.~Kumar, S.~Ermon, and B.~Poole, ``{Score-Based Generative Modeling through Stochastic Differential Equations},'' in \emph{Proc. Int. Conf. Learn. Represent.}, 2021.

\bibitem{Avrahami2021BlendedDF}
O.~Avrahami, D.~Lischinski, and O.~Fried, ``{Blended Diffusion for Text-driven Editing of Natural Images},'' in \emph{Proc. IEEE Conf. Comput. Vis. Pattern Recognit.}, 2022, pp. 18\,208--18\,218.

\bibitem{Couairon2022DiffEditDS}
G.~Couairon, J.~Verbeek, H.~Schwenk, and M.~Cord, ``{DiffEdit: Diffusion-based semantic image editing with mask guidance},'' \emph{arXiv}, vol. abs/2210.11427, 2022.

\bibitem{Hertz2022PrompttoPromptIE}
A.~Hertz, R.~Mokady, J.~M. Tenenbaum, K.~Aberman, Y.~Pritch, and D.~Cohen-Or, ``Prompt-to-prompt image editing with cross attention control,'' \emph{arXiv}, vol. abs/2208.01626, 2022.

\bibitem{Meng2021SDEditGI}
C.~Meng, Y.~He, Y.~Song, J.~Song, J.~Wu, J.-Y. Zhu, and S.~Ermon, ``{{SDE}dit: Guided Image Synthesis and Editing with Stochastic Differential Equations},'' in \emph{Proc. Int. Conf. Learn. Represent.}, 2022.

\bibitem{Mokady2022NulltextIF}
R.~Mokady, A.~Hertz, K.~Aberman, Y.~Pritch, and D.~Cohen-Or, ``{Null-text Inversion for Editing Real Images using Guided Diffusion Models},'' in \emph{Proc. IEEE Conf. Comput. Vis. Pattern Recognit.}, 2023, pp. 6038--6047.

\bibitem{Rombach2021HighResolutionIS}
R.~Rombach, A.~Blattmann, D.~Lorenz, P.~Esser, and B.~Ommer, ``{High-Resolution Image Synthesis with Latent Diffusion Models},'' in \emph{Proc. IEEE Conf. Comput. Vis. Pattern Recognit.}, 2022, pp. 10\,684--10\,695.

\bibitem{Saharia2022PhotorealisticTD}
C.~Saharia, W.~Chan, S.~Saxena, L.~Li, J.~Whang, E.~L. Denton, S.~K.~S. Ghasemipour, B.~K. Ayan, R.~G. Lopes, T.~Salimans, J.~Ho, D.~J. Fleet, and M.~Norouzi, ``{Photorealistic Text-to-Image Diffusion Models with Deep Language Understanding},'' in \emph{Proc. Adv. Neural Inf. Process. Syst.}, 2022.

\bibitem{Wallace2022EDICTED}
B.~Wallace, A.~Gokul, and N.~V. Naik, ``{{EDICT}: Exact Diffusion Inversion via Coupled Transformations},'' in \emph{Proc. IEEE Conf. Comput. Vis. Pattern Recognit.}, 2023, pp. 22\,532--22\,541.

\bibitem{Xie2022SmartBrushTA}
S.~Xie, Z.~Zhang, Z.~Lin, T.~Hinz, and K.~Zhang, ``{SmartBrush: Text and Shape Guided Object Inpainting with Diffusion Model},'' in \emph{Proc. IEEE Conf. Comput. Vis. Pattern Recognit.}, 2023, pp. 22\,428--22\,437.

\bibitem{Xu2023OpenVocabularyPS}
J.~Xu, S.~Liu, A.~Vahdat, W.~Byeon, X.~Wang, and S.~D. Mello, ``{Open-Vocabulary Panoptic Segmentation with Text-to-Image Diffusion Models},'' in \emph{Proc. IEEE Conf. Comput. Vis. Pattern Recognit.}, 2023, pp. 2955--2966.

\bibitem{Zhu2023MovieFactoryAM}
J.~Zhu, H.~Yang, H.~He, W.~Wang, Z.~Tuo, W.-H. Cheng, L.~Gao, J.~Song, and J.~Fu, ``{MovieFactory: Automatic Movie Creation from Text using Large Generative Models for Language and Images},'' in \emph{Proc. 31st ACM Int. Conf. Multimedia}, 2023.

\bibitem{Tu2023MotionEditorEV}
S.~Tu, Q.~Dai, Z.-Q. Cheng, H.-R. Hu, X.~Han, Z.~Wu, and Y.-G. Jiang, ``{MotionEditor: Editing Video Motion via Content-Aware Diffusion},'' in \emph{Proc. IEEE/CVF Conf. Comput. Vis. Pattern Recog.}, 2023.

\bibitem{Tu2024StableAnimatorHI}
S.~Tu, Z.~Xing, X.~Han, Z.-Q. Cheng, Q.~Dai, C.~Luo, and Z.~Wu, ``{StableAnimator: High-Quality Identity-Preserving Human Image Animation},'' in \emph{Proc. IEEE/CVF Conf. Comput. Vis. Pattern Recog.}, 2024.

\bibitem{Maras2018DeterminingAO}
M.-H. Maras and A.~Alexandrou, ``Determining authenticity of video evidence in the age of artificial intelligence and in the wake of deepfake videos,'' \emph{Int. J. Evid. Proof}, vol.~23, pp. 255 -- 262, 2018.

\bibitem{Pei2024DeepfakeGA}
G.~Pei, J.~Zhang, M.~Hu, G.~Zhai, C.~Wang, Z.~Zhang, J.~Yang, C.~Shen, and D.~Tao, ``Deepfake generation and detection: A benchmark and survey,'' \emph{arXiv}, vol. abs/2403.17881, 2024.

\bibitem{Passos2022ARO}
L.~A. Passos, D.~S. Jodas, K.~A.~P. Costa, L.~A.~S. J'unior, D.~Colombo, and J.~P. Papa, ``A review of deep learning‐based approaches for deepfake content detection,'' \emph{Expert Syst.}, vol.~41, 2022.

\bibitem{Naitali2023DeepfakeAG}
A.~Naitali, M.~Ridouani, F.~Salahdine, and N.~Kaabouch, ``Deepfake attacks: Generation, detection, datasets, challenges, and research directions,'' \emph{Comput.}, vol.~12, p. 216, 2023.

\bibitem{wang2022lisiam}
J.~Wang, Y.~Sun, and J.~Tang, ``Lisiam: Localization invariance siamese network for deepfake detection,'' \emph{IEEE Trans. Inf. Forensics Security}, vol.~17, pp. 2425--2436, 2022.

\bibitem{han2023fcd}
R.~Han, X.~Wang, N.~Bai, Q.~Wang, Z.~Liu, and J.~Xue, ``Fcd-net: Learning to detect multiple types of homologous deepfake face images,'' \emph{IEEE Trans. Inf. Forensics Security}, vol.~18, pp. 2653--2666, 2023.

\bibitem{yin2023dynamic}
Q.~Yin, W.~Lu, B.~Li, and J.~Huang, ``Dynamic difference learning with spatio--temporal correlation for deepfake video detection,'' \emph{IEEE Trans. Inf. Forensics Security}, vol.~18, pp. 4046--4058, 2023.

\bibitem{10061274}
G.-L. Chen and C.-C. Hsu, ``Jointly defending deepfake manipulation and adversarial attack using decoy mechanism,'' \emph{IEEE Trans. Pattern Anal. Mach. Intell.}, vol.~45, no.~8, pp. 9922--9931, 2023.

\bibitem{10411047}
T.~Qiao, S.~Xie, Y.~Chen, F.~Retraint, and X.~Luo, ``Fully unsupervised deepfake video detection via enhanced contrastive learning,'' \emph{IEEE Trans. Pattern Anal. Mach. Intell.}, vol.~46, no.~7, pp. 4654--4668, 2024.

\bibitem{9468380}
Y.~Nirkin, L.~Wolf, Y.~Keller, and T.~Hassner, ``Deepfake detection based on discrepancies between faces and their context,'' \emph{IEEE Trans. Pattern Anal. Mach. Intell.}, vol.~44, no.~10, pp. 6111--6121, 2022.

\bibitem{10209264}
Q.~Ying, H.~Zhou, Z.~Qian, S.~Li, and X.~Zhang, ``Learning to immunize images for tamper localization and self-recovery,'' \emph{IEEE Trans. Pattern Anal. Mach. Intell.}, vol.~45, no.~11, pp. 13\,814--13\,830, 2023.

\bibitem{Aneja2021TAFIMTA}
S.~Aneja, L.~Markhasin, and M.~Nie{\ss}ner, ``{TAFIM: Targeted Adversarial Attacks against Facial Image Manipulations},'' in \emph{Proc. Eur. Conf. Comput. Vis.}, 2022.

\bibitem{Ruiz2020DisruptingDA}
N.~Ruiz, S.~A. Bargal, and S.~Sclaroff, ``{Disrupting Deepfakes: Adversarial Attacks Against Conditional Image Translation Networks and Facial Manipulation Systems},'' \emph{arXiv}, vol. abs/2003.01279, 2020.

\bibitem{Ruiz2023PracticalDO}
N.~Ruiz, S.~A. Bargal, C.~Xie, and S.~Sclaroff, ``Practical disruption of image translation deepfake networks,'' in \emph{Proc. AAAI Conf. Artif. Intell.}, 2023.

\bibitem{Yeh2021AttackAT}
C.-Y. Yeh, H.-W. Chen, H.-H. Shuai, D.-N. Yang, and M.-S. Chen, ``Attack as the best defense: Nullifying image-to-image translation gans via limit-aware adversarial attack,'' in \emph{Proc. IEEE Int. Conf. Comput. Vis.}, 2021, pp. 16\,168--16\,177.

\bibitem{Salman2023RaisingTC}
H.~Salman, A.~Khaddaj, G.~Leclerc, A.~Ilyas, and A.~Madry, ``{Raising the Cost of Malicious AI-Powered Image Editing},'' in \emph{Proc. Int. Conf. Mach. Learn.}, 2023.

\bibitem{Xue2023TowardEP}
H.~Xue, C.~Liang, X.~Wu, and Y.~Chen, ``{Toward Effective Protection Against Diffusion Based Mimicry Through Score Distillation},'' \emph{arXiv}, vol. abs/2311.12832, 2023.

\bibitem{Lo_2024_CVPR}
L.~Lo, C.~Y. Yeo, H.-H. Shuai, and W.-H. Cheng, ``{Distraction is All You Need: Memory-Efficient Image Immunization against Diffusion-Based Image Editing},'' in \emph{Proc. IEEE Conf. Comput. Vis. Pattern Recognit.}, 2024, pp. 24\,462--24\,471.

\bibitem{Liang2023MistTI}
C.~Liang and X.~Wu, ``{Mist: Towards Improved Adversarial Examples for Diffusion Models},'' \emph{arXiv}, vol. abs/2305.12683, 2023.

\bibitem{Zheng2023TargetedAI}
B.~Zheng, C.~Liang, X.~Wu, and Y.~Liu, ``{Targeted Attack Improves Protection against Unauthorized Diffusion Customization},'' \emph{arXiv}, vol. abs/2310.04687, 2023.

\bibitem{Fan2023TransferabilityBT}
M.~Fan, X.~Li, C.~Chen, W.~Zhou, and Y.~Li, ``Transferability bound theory: Exploring relationship between adversarial transferability and flatness,'' in \emph{Proc. Adv. Neural Inf. Process. Syst.}, 2023.

\bibitem{Luo2024AnII}
H.~Luo, J.~Gu, F.~Liu, and P.~Torr, ``An image is worth 1000 lies: Adversarial transferability across prompts on vision-language models,'' \emph{arXiv}, vol. abs/2403.09766, 2024.

\bibitem{meng2021sdedit}
C.~Meng, Y.~Song, J.~Song, J.~Wu, and S.~Ermon, ``{{SDE}dit: Image Synthesis and Editing with Stochastic Differential Equations},'' \emph{arXiv}, vol. abs/2108.01073, 2021.

\bibitem{avrahami2022blended}
O.~Avrahami, O.~Fried, and D.~Cohen-Or, ``{Blended Diffusion for Text-Driven Editing of Natural Images},'' \emph{arXiv}, vol. abs/2111.14818, 2022.

\bibitem{Radford2021LearningTV}
A.~Radford, J.~W. Kim, C.~Hallacy, A.~Ramesh, G.~Goh, S.~Agarwal, G.~Sastry, A.~Askell, P.~Mishkin, J.~Clark, G.~Krueger, and I.~Sutskever, ``Learning transferable visual models from natural language supervision,'' in \emph{Proc. Int. Conf. Mach. Learn.}, 2021.

\bibitem{kim2021diffusionclip}
G.~Kim and Y.~Choi, ``{{DiffusionCLIP}: Text-Guided Image Manipulation Using Diffusion Models},'' \emph{arXiv}, vol. abs/2110.02711, 2021.

\bibitem{hertz2022prompt}
A.~Hertz, R.~Mokady, J.~Tenenbaum, K.~Aberman, Y.~Pritch, and D.~Cohen-Or, ``{Prompt-to-Prompt Image Editing with Cross Attention Control},'' \emph{arXiv}, vol. abs/2208.01626, 2022.

\bibitem{zhang2023controlnet}
L.~Zhang and M.~Agrawala, ``{ControlNet: Adding Conditional Control to Text-to-Image Diffusion Models},'' \emph{arXiv}, vol. abs/2302.05543, 2023.

\bibitem{Shan2020FawkesPP}
S.~Shan, E.~Wenger, J.~Zhang, H.~Li, H.~Zheng, and B.~Y. Zhao, ``Fawkes: Protecting personal privacy against unauthorized deep learning models,'' \emph{arXiv}, vol. abs/2002.08327, 2020.

\bibitem{Dong2017DiscoveringAE}
Y.~Dong, F.~Liao, T.~Pang, X.~Hu, and J.~Zhu, ``Discovering adversarial examples with momentum,'' \emph{arXiv}, vol. abs/1710.06081, 2017.

\bibitem{MoosaviDezfooli2015DeepFoolAS}
S.-M. Moosavi-Dezfooli, A.~Fawzi, and P.~Frossard, ``Deepfool: A simple and accurate method to fool deep neural networks,'' in \emph{Proc. IEEE Conf. Comput. Vis. Pattern Recognit.}, 2016, pp. 2574--2582.

\bibitem{Xie2017AdversarialEF}
C.~Xie, J.~Wang, Z.~Zhang, Y.~Zhou, L.~Xie, and A.~L. Yuille, ``Adversarial examples for semantic segmentation and object detection,'' in \emph{Proc. IEEE Int. Conf. Comput. Vis.}, 2017, pp. 1378--1387.

\bibitem{Wang2021AdmixET}
X.~Wang, X.~He, J.~Wang, and K.~He, ``Admix: Enhancing the transferability of adversarial attacks,'' in \emph{Proc. IEEE Int. Conf. Comput. Vis.}, 2021, pp. 16\,138--16\,147.

\bibitem{Xie2018ImprovingTO}
C.~Xie, Z.~Zhang, J.~Wang, Y.~Zhou, Z.~Ren, and A.~L. Yuille, ``Improving transferability of adversarial examples with input diversity,'' in \emph{Proc. IEEE Conf. Comput. Vis. Pattern Recognit.}, 2019, pp. 2725--2734.

\bibitem{Lin2019NesterovAG}
J.~Lin, C.~Song, K.~He, L.~Wang, and J.~E. Hopcroft, ``Nesterov accelerated gradient and scale invariance for adversarial attacks,'' \emph{arXiv}, vol. abs/1908.06281, 2019.

\bibitem{Dong2017BoostingAA}
Y.~Dong, F.~Liao, T.~Pang, H.~Su, J.~Zhu, X.~Hu, and J.~Li, ``Boosting adversarial attacks with momentum,'' in \emph{Proc. IEEE Conf. Comput. Vis. Pattern Recognit.}, 2018, pp. 9185--9193.

\bibitem{NEURIPS2022_c0f9419c}
Z.~Qin, Y.~Fan, Y.~Liu, L.~Shen, Y.~Zhang, J.~Wang, and B.~Wu, ``Boosting the transferability of adversarial attacks with reverse adversarial perturbation,'' in \emph{Proc. Adv. Neural Inf. Process. Syst.}, 2022, pp. 29\,845--29\,858.

\bibitem{Goodfellow2014ExplainingAH}
I.~J. Goodfellow, J.~Shlens, and C.~Szegedy, ``Explaining and harnessing adversarial examples,'' \emph{arXiv}, vol. abs/1412.6572, 2014.

\bibitem{Kurakin2016AdversarialEI}
A.~Kurakin, I.~J. Goodfellow, and S.~Bengio, ``Adversarial examples in the physical world,'' \emph{arXiv}, vol. abs/1607.02533, 2016.

\bibitem{Madry2017TowardsDL}
A.~Madry, A.~Makelov, L.~Schmidt, D.~Tsipras, and A.~Vladu, ``Towards deep learning models resistant to adversarial attacks,'' \emph{arXiv}, vol. abs/1706.06083, 2017.

\bibitem{Foret2020SharpnessAwareMF}
P.~Foret, A.~Kleiner, H.~Mobahi, and B.~Neyshabur, ``Sharpness-aware minimization for efficiently improving generalization,'' \emph{arXiv}, vol. abs/2010.01412, 2020.

\bibitem{Esser2024ScalingRF}
P.~Esser, S.~Kulal, A.~Blattmann, R.~Entezari, J.~Muller, H.~Saini, Y.~Levi, D.~Lorenz, A.~Sauer, F.~Boesel, D.~Podell, T.~Dockhorn, Z.~English, K.~Lacey, A.~Goodwin, Y.~Marek, and R.~Rombach, ``Scaling rectified flow transformers for high-resolution image synthesis,'' \emph{arXiv}, vol. abs/2403.03206, 2024.

\bibitem{Brooks2022InstructPix2PixLT}
T.~Brooks, A.~Holynski, and A.~A. Efros, ``{InstructPix2Pix: Learning to Follow Image Editing Instructions},'' in \emph{Proc. IEEE Conf. Comput. Vis. Pattern Recognit.}, 2023, pp. 18\,392--18\,402.

\bibitem{huggingface}
H.~Face, ``Hugging face: The ai community building the future,'' \url{https://huggingface.co/}, 2025.

\bibitem{instructpix2pix-clip-filtered}
T.~Brooks, A.~Holynski, and A.~A. Efros, ``instructpix2pix-clip-filtered,'' \url{https://huggingface.co/datasets/timbrooks/instructpix2pix-clip-filtered}, 2022.

\bibitem{1284395}
Z.~Wang, A.~C. Bovik, H.~R. Sheikh, and E.~P. Simoncelli, ``{Image quality assessment: from error visibility to structural similarity},'' \emph{IEEE Trans. Image Process.}, vol.~13, no.~4, pp. 600--612, 2004.

\bibitem{1576816}
H.~R. Sheikh and A.~C. Bovik, ``{Image information and visual quality},'' \emph{IEEE Trans. Image Process.}, vol.~15, no.~2, pp. 430--444, 2006.

\bibitem{5705575}
L.~Zhang, L.~Zhang, X.~Mou, and D.~Zhang, ``{{FSIM}: A Feature Similarity Index for Image Quality Assessment},'' \emph{IEEE Trans. Image Process.}, vol.~20, no.~8, pp. 2378--2386, 2011.

\bibitem{Zhang2018TheUE}
R.~Zhang, P.~Isola, A.~A. Efros, E.~Shechtman, and O.~Wang, ``{The Unreasonable Effectiveness of Deep Features as a Perceptual Metric},'' in \emph{Proc. IEEE Conf. Comput. Vis. Pattern Recognit.}, 2018, pp. 586--595.

\end{thebibliography}

\vspace{-22pt}
\begin{IEEEbiography}[{\includegraphics[width=1in,height=1.25in,clip,keepaspectratio]{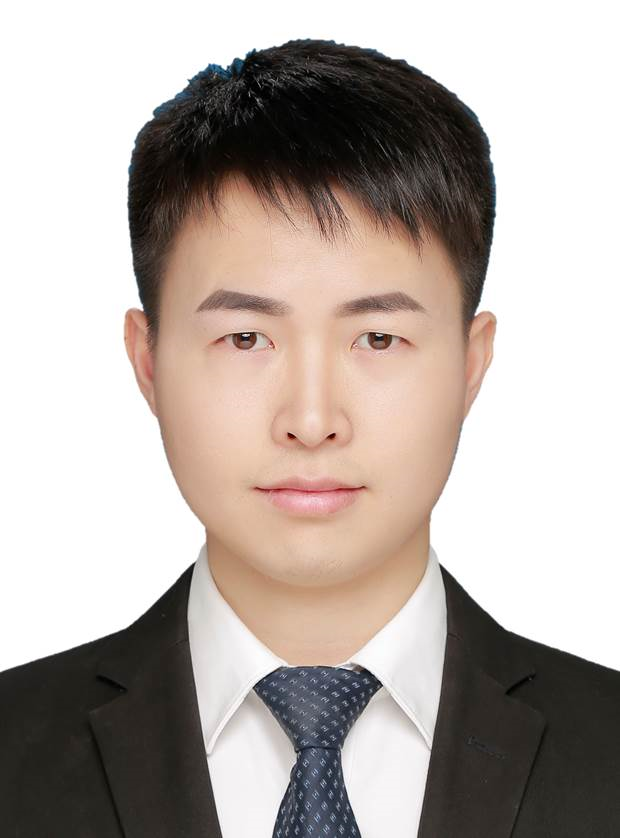}}]{Jie Zhang}
(Member, IEEE) received the Ph.D. degree from the University of Chinese Academy of Sciences (CAS), Beijing, China. He is currently an Associate Professor with the Institute of Computing Technology, CAS. His research interests include computer vision, pattern recognition, machine learning, particularly include adversarial attacks and defenses, domain generalization, AI safety and trustworthiness.
\end{IEEEbiography}
\vspace{-22pt}

\begin{IEEEbiography}[{\includegraphics[width=1in,height=1.25in,clip,keepaspectratio]{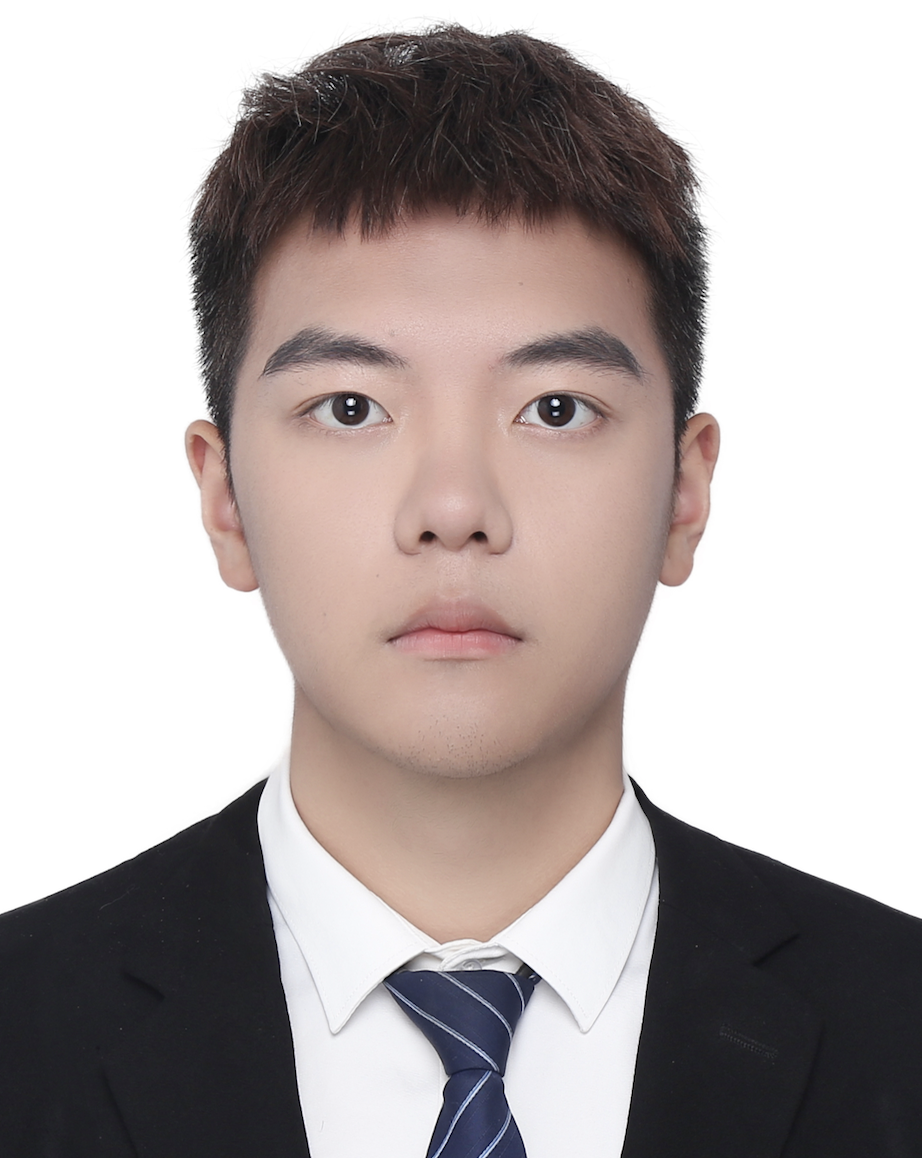}}]{Shuai Dong} received the B.Sc. degree (expected) in intelligent science and technology from China University of Geosciences, Wuhan, China. His research interests include computer vision, pattern recognition, machine learning, particularly include adversarial attacks and defenses, AI safety and trustworthiness.
\end{IEEEbiography}
\vspace{-22pt}

\begin{IEEEbiography}[{\includegraphics[width=1in,height=1.25in,clip,keepaspectratio]{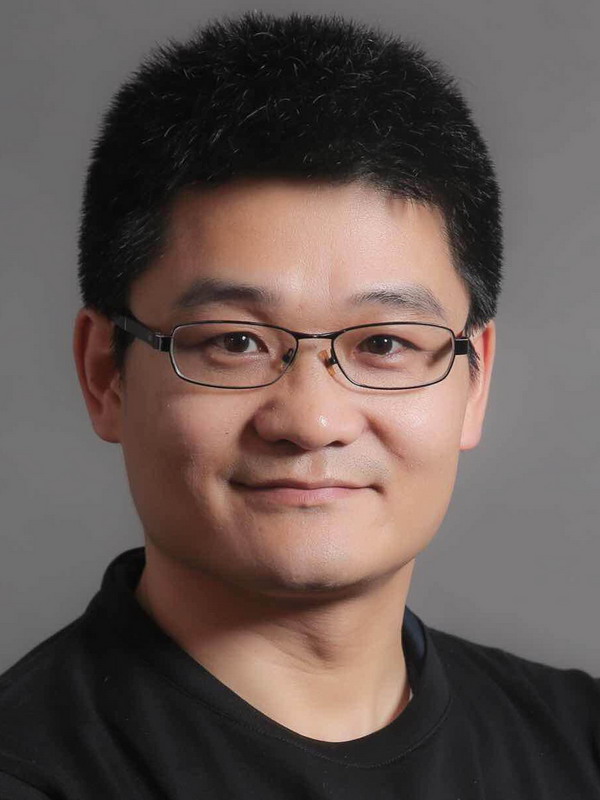}}]{Shiguang Shan}
(Fellow, IEEE) received the Ph.D. degree in computer science from the Institute of Computing Technology (ICT), Chinese Academy of Sciences (CAS), Beijing, China, in 2004. He has been a Full Professor with ICT since 2010, where he is currently the Director of the Key Laboratory of Intelligent Information Processing, CAS. His research interests include signal processing, computer vision, pattern recognition, and machine learning. He has published more than 300 articles in related areas. He served as the General Co-Chair for IEEE Face and Gesture Recognition 2023, the General Co-Chair for Asian Conference on Computer Vision (ACCV) 2022, and the Area Chair of many international conferences, including CVPR, ICCV, AAAI, IJCAI, ACCV, ICPR, and FG. He was/is an Associate Editors of several journals, including IEEE Transactions on Image Processing, Neurocomputing, CVIU, and PRL. He was a recipient of the China's State Natural Science Award in 2015 and the China’s State S\&T Progress Award in 2005 for his research work.
\end{IEEEbiography}
\vspace{-22pt}

\begin{IEEEbiography}[{\includegraphics[width=1in,height=1.25in,clip,keepaspectratio]{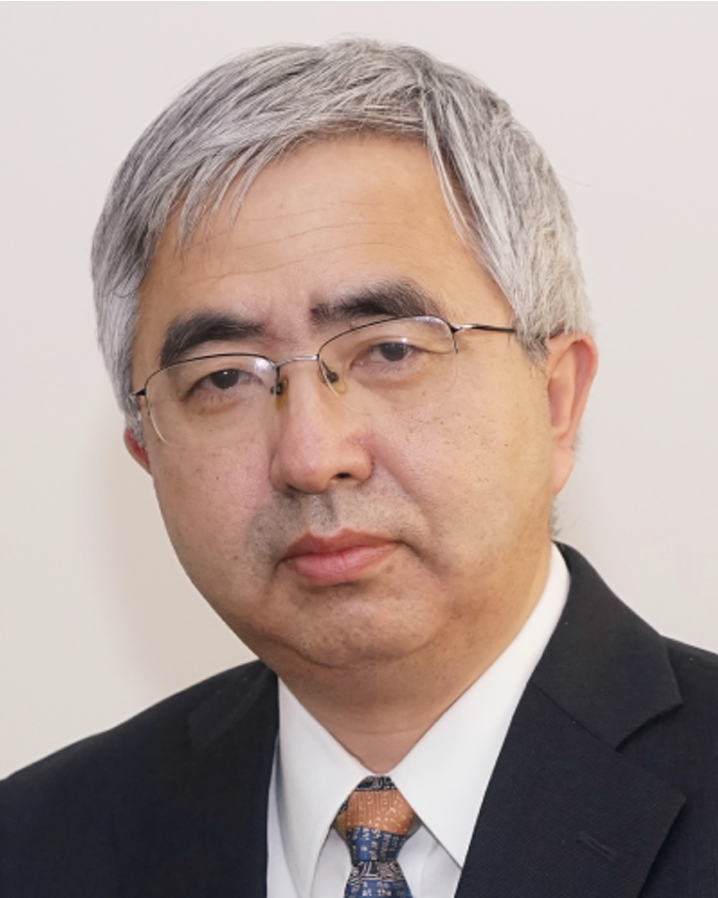}}]{Xilin Chen} (Fellow, IEEE) is currently a Professor with the Institute of Computing Technology, Chinese
 Academy of Sciences (CAS). He has authored one
 book and more than 400 articles in refereed journals
 and proceedings in the areas of computer vision,
 pattern recognition, image processing, and multi
modal interfaces. He is a fellow of the ACM,
 IAPR, and CCF. He is also an Information Sciences
 Editorial Board Member of Fundamental Research,
 an Editorial Board Member of Research, a Senior
 Editor of the Journal of Visual Communication and
 Image Representation, and an Associate Editor-in-Chief of the Chinese Jour
nal of Computers and Chinese Journal of Pattern Recognition and Artificial
 Intelligence. He served as an organizing committee member for multiple
 conferences, including the General Co-Chair of FG 2013/FG 2018, VCIP
 2022, the Program Co-Chair of ICMI 2010/FG 2024, and an Area Chair of
 ICCV/CVPR/ECCV/NeurIPS for more than ten times.
\end{IEEEbiography}



\vfill

\end{document}